\documentclass{article}
\usepackage[accepted]{icml2024}
\usepackage{microtype}
\usepackage{graphicx}
\usepackage{booktabs} 
\usepackage{graphicx}
\usepackage{subcaption}
\usepackage{caption}
\usepackage{hyperref}

\usepackage{icml2024}

\usepackage{enumitem}
\usepackage{bm}
\usepackage{bbm}
\usepackage{comment}
\usepackage{amsmath}
\usepackage{amssymb}
\usepackage{mathtools}
\usepackage{amsthm}

\usepackage[capitalize,noabbrev]{cleveref}

\theoremstyle{plain}
\newtheorem{theorem}{Theorem}[section]

\newtheorem{lemma}[theorem]{Lemma}

\theoremstyle{definition}
\newtheorem{definition}[theorem]{Definition}
\newtheorem{assumption}[theorem]{Assumption}

\theoremstyle{definition}
\newtheorem{remark}{Remark}

\usepackage[textsize=tiny]{todonotes}

\icmltitlerunning{Understanding Forgetting in Continual Learning with Linear Regression}

\begin{document}

\twocolumn[
\icmltitle{Understanding Forgetting in Continual Learning with Linear Regression:\\ Overparameterized and Underparameterized Regimes}




\begin{icmlauthorlist}
\icmlauthor{Meng Ding}{ub}
\icmlauthor{Kaiyi Ji}{ub}
\icmlauthor{Di Wang}{sch}
\icmlauthor{Jinhui Xu}{ub}
\end{icmlauthorlist}

\icmlaffiliation{ub}{Department of Computer Science and Engineering, State University of New York at Buffalo, Buffalo, USA}
\icmlaffiliation{sch}{Division of CEMSE, King Abdullah University of Science and Technology, Thuwal, Saudi Arabia}

\icmlcorrespondingauthor{Kaiyi Ji}{kaiyiji@buffalo.edu}
\icmlcorrespondingauthor{Jinhui Xu}{jinhui@buffalo.edu}

\icmlkeywords{Machine Learning, ICML}

\vskip 0.3in
]



\printAffiliationsAndNotice{} 
\begin{abstract}

Continual learning, focused on sequentially learning multiple tasks, has gained significant attention recently. Despite the tremendous progress made in the past, the theoretical understanding, especially factors contributing to \emph{catastrophic forgetting}, remains relatively unexplored. In this paper, we provide a general theoretical analysis of forgetting in the linear regression model via Stochastic Gradient Descent (SGD) applicable to both under-parameterized and overparameterized regimes. Our theoretical framework reveals some interesting insights into the intricate relationship between task sequence and algorithmic parameters, an aspect not fully captured in previous studies due to their restrictive assumptions. Specifically, {we demonstrate that, given a sufficiently large data size, the arrangement of tasks in a sequence—where tasks with larger eigenvalues in their population data covariance matrices are trained later—tends to result in increased forgetting.} Additionally, our findings highlight that an appropriate choice of step size will help mitigate forgetting in both under-parameterized and overparameterized settings.
To validate our theoretical analysis, we conducted simulation experiments on both linear regression models and Deep Neural Networks (DNNs). Results from these simulations substantiate our theoretical findings.
\end{abstract}

\section{Introduction}
Continual learning, also known as lifelong learning, is a subfield of machine learning that focuses on developing a model capable of learning continuously from a stream of data, which are i.i.d sampled from different tasks and presented sequentially to the model. A primary challenge in continual learning is the \emph{catastrophic forgetting} phenomenon \cite{mccloskey1989catastrophic}, wherein the model forgets previously acquired knowledge when exposed to new data.

Previous research addressing catastrophic forgetting in continuous learning primarily focuses on empirical studies, which can be broadly classified into three categories: expansion-based methods, regularization-based methods, and memory-based methods. Expansion-based methods \cite{yoon2017lifelong, yoon2019scalable, yang2021grown} mitigate catastrophic forgetting by allocating distinct subsets of network parameters to individual tasks. Regularization-based methods \cite{kirkpatrick2017overcoming, aljundi2018memory, serra2018overcoming, liu2022continual} employee structural regularization in fixed capacity models to counteract forgetting, which penalize significant changes in parameters that are crucial for previous tasks. Memory-based methods \cite{shin2017continual,chaudhry2018efficient, riemer2018learning, saha2021gradient, lin2022trgp,hao2023bilevel} alleviate forgetting by storing subsets of previous task data or synthesizing pseudo-data without data-replay.

Recently, there has been a growing body of work focused on understanding the behavior of catastrophic forgetting from a theoretical standpoint. For example, \citealt{bennani2020generalisation, doan2021theoretical} analyze the generalization of continual learning for Orthogonal Gradient Descent (OGD) \cite{farajtabar2020orthogonal} in the  Neural Tangent Kernel (NTK) \cite{jacot2018neural} regime. \citealt{lee2021continual, asanuma2021statistical} explore the impact of task similarity in a teacher-student setting. \citealt{evron2022catastrophic, lin2023theory} provide a detailed forgetting analysis of the \emph{minimum-norm interpolator} for the overparameterized linear regression model. 
However, the existing analyses of forgetting often rely on relatively stringent assumptions that may not be applicable in many scenarios. For example, \citealt{bennani2020generalisation,doan2021theoretical,evron2022catastrophic,lin2023theory} necessitate an overparameterized regime for their analysis, which may be invalid when involving large datasets. Moreover, \citealt{lee2021continual, asanuma2021statistical,lin2023theory,swartworth2023nearly} assume that data follows a Gaussian distribution that may not hold in real-world datasets exhibiting more complex distributions. \citealt{evron2022catastrophic,lin2023theory} focus on the minimum-norm interpolator, where each task requires achieving zero loss on its training samples and hence can find a closed-form solution.

In this paper, we investigate the behavior of forgetting under the linear regression model via the more practical Stochastic Gradient Descent (SGD) method and provide a general theoretical analysis that is applicable to both over-parameterized and under-parameterized regimes. Our main contributions can be summarized as follows:

Firstly, our work provides a theoretical analysis for multi-step SGD algorithms in both underparameterized and overparameterized regimes, with the population data covariance matrix satisfying the general fourth moment instead of Gaussian distribution as in existing studies. In specific, we provide a novel upper bound on the model forgetting, as well as a matching lower bound that shows the tightness of our characterization. Our bounds derive the forgetting bound that is stated as a function of $\mathbf{1)}$ the spectrum of the population data covariance matrices for each task, $\mathbf{2)}$ the step size, $\mathbf{3)}$ the number of training samples and $\mathbf{4)}$ the effective dimensions on the forgetting.  

Second, our study provides some interesting insights into the impact of task sequence and algorithmic parameters on the degree of forgetting.  
Specifically, we show that when the data size is sufficiently large, forgetting tends to escalate when we postpone the training of tasks, whose population data covariance matrices possess larger eigenvalues. 
It is intuitive that when tasks with larger eigenvalues are trained later, the model might overfit these tasks due to their high variance. In addition, our findings reveal that an appropriate choice of step size can help mitigate forgetting in both underparameterized and overparameterized settings. Note that these results cannot be derived from existing works due to their restrictive data distribution assumptions or closed-form updating rules. More detailed discussions can be found in \cref{sec:discuss}.

{Finally, we conducted simulation experiments on both linear regression models and Deep Neural Networks (DNNs) to validate our theoretical analysis. Our simulation results indicate that both linear regression models and DNNs exhibit increased forgetting when tasks with larger eigenvalues are encountered later.} Additionally, we demonstrate that smaller step sizes in training can also mitigate forgetting across task sequences, especially in under-parameterized settings. Interestingly, we observe that in over-parameterized DNNs, higher dimensionality does not necessarily equate to more forgetting if the dataset size is fixed, as opposite to the linear regression case.
\vspace{-5pt}
\subsection{Related Work}
In this section, we discuss related work on Covariate Shift, SGD analysis in linear regression, and theoretical studies for catastrophic forgetting.

\textbf{Covariate Shift}
Covariate shift is a specific set-up in machine learning \cite{pan2009survey,sugiyama2012machine}, referring to a distribution mismatch between the training and test data. The concept is typically applied in transfer learning, which can be seen as a particular instance of continual learning, generally involving two tasks. For example, \citealt{mohri2012new,cortes2014domain,kpotufe2018marginal,cortes2019adaptation,hanneke2020value,ma2023optimally,wu2022power} examine the (regularized) empirical risk minimizer, which focuses on minimizing the empirical and generalization error across accessible datasets. Nevertheless, the standard covariate shift is defined over 
two distinct data distributions, which can not be directly applied to our case. Consequently, we propose an extended version in \cref{def:covariate shift} to better suit our context.

\textbf{SGD Analysis}
Recently, several studies have investigated the behavior of Stochastic Gradient Descent (SGD) in linear regression models through the lens of bias-variance decomposition \cite{defossez2015averaged, dieuleveut2017harder, jain2017markov, jain2018parallelizing} and the eigen-decomposition of the covariance matrix \cite{chen2020dimension,zou2021benign,wu2022last,wu2022power}. Our work closely relates to the studies in \citealt{zou2021benign,wu2022power} that also characterized the SGD dynamic in linear regression with respect to the full eigenspectrum of the data covariance matrix. However, they focused on either the single-task setting or the pretraining-finetuning setting, while we studied the more challenging continual learning problem that involves a sequence of tasks with different data distributions.  More discussion in \cref{sec:discuss}.

\textbf{Theoretical Studies in Continual Learning}
Although significant progress has been made in empirical studies addressing the issue of forgetting in continual learning, theoretical insights into this area are still largely unexplored. In this context, \citealt{bennani2020generalisation} established a theoretical framework to study continual learning algorithms in the NTK regime, and provided the first generalization bound dependent on task similarity for SGD and OGD. \citealt{doan2021theoretical} introduced the NTK overlap matrix as a task similarity metric and proposed a data-structure-informed variant of OGD that utilizes Principal Component Analysis (PCA). \citealt{asanuma2021statistical} utilized the teacher-student framework on a single neural network and demonstrated that catastrophic forgetting can be circumvented when the similarity among input distributions is small and the similarity among teacher networks is large.  \citealt{lee2021continual} expanded an earlier analysis of two-layer networks within the teacher-student setup to the setting with multiple teachers and revealed that the highest level of forgetting occurs when tasks have intermediate similarity with each other. \citealt{evron2022catastrophic,swartworth2023nearly} explained the behavior of forgetting in the linear regression model from the perspectives of alternating projections and the Kaczmarz method \cite{karczmarz1937angenaherte}. \citealt{lin2023theory} investigated the impact of overparameterization, task similarity, and task ordering on forgetting and generalization in the overparameterized linear regression model.

The works most relevant to our study include \cite{evron2022catastrophic, lin2023theory}, both of which also studied the behavior of forgetting in the linear regression model. However, our work differs from their studies in several aspects.

Firstly, with regard to assumptions, \citealt{evron2022catastrophic} assumed all data are bounded with 1 and the model is noiseless, and \citealt{lin2023theory} assumed all data are sampled from a Gaussian distribution. In contrast, our assumptions cover more data distributions and are much milder than theirs (see Remark 2 and Section 4 for more details). Secondly, in terms of methods, both \citealt{evron2022catastrophic} and \citealt{lin2023theory} analyze the problem of forgetting using the minimum norm solution, which presupposes zero training error—a requirement not necessary in our approach with SGD (see Section 2 for further discussions). Third, \citealt{evron2022catastrophic,lin2023theory} considered only the overparameterized case where the data dimension is larger than the data size, while our analysis holds for both the underparameterized and overparameterized settings. 

\textbf{Notations}: {In this paper, we adhere to a consistent notation style for clarity.} We use boldface lower letters such as $\mathbf{x}, \mathbf{w}$ for vectors, and boldface capital letters (e.g. $\mathbf{A}, \mathbf{H}$) for matrices. Let $\|\mathbf{A}\|_2$ denote the spectral norm of $\mathbf{A}$ and $\|\mathbf{v}\|_2$ denote the Euclidean norm of $\mathbf{v}$. For two vectors $\mathbf{u}$ and $\mathbf{v}$, their inner product is denoted by $\langle\mathbf{u}, \mathbf{v}\rangle$ or $\mathbf{u}^{\top} \mathbf{v}$. For two matrices $\mathbf{A}$ and $\mathbf{B}$ of appropriate dimension, their inner product is defined as $\langle\mathbf{A}, \mathbf{B}\rangle:=\operatorname{tr} (\mathbf{A}^{\top} \mathbf{B} )$. For a positive semi-definite (PSD) matrix $\mathbf{A}$ and a vector $\mathbf{v}$ of appropriate dimension, we write $\|\mathbf{v}\|_{\mathbf{A}}^2:=\mathbf{v}^{\top} \mathbf{A v}$. The outer product is denoted by $\otimes$. 

\section{Preliminaries}
{In our setup, we consider a sequence of tasks, denoted as $\mathbb{M}=\{1,2, \ldots, M\}$. For each task $m$ in this sequence, we have a corresponding dataset $D_m$, which consists of $N$ data points. Each of these data points, denoted as $(\mathbf{x}_{m, i}, y_{m, i})$, is drawn independently and identically distributed (i.i.d.) from a specific distribution $\mathcal{D}_m=$ $\mathcal{X}_m \times \mathcal{Y}_m \subset \mathbb{R}^d \times \mathbb{R}$. Here, $\mathbf{x}_{m, i}$ represents the feature vector, and $y_{m, i}$ is the response variable for each data point in the dataset $D_m$.}
Assume that $ \{(\mathbf{x}_{m,i}, y_{m,i}) \}_{i=1}^N$ are i.i.d. sampled from a linear regression model, i.e., each pair $(\mathbf{x}_{m,i}, y_{m,i})$ is a realization of the linear regression model $ y_m  =  (\mathbf{x}_m^{\top} \mathbf{w}_* ) + z_m$, where $z_m$ is some randomized noise and $\mathbf{w}_*\in \mathbb{R}^d$ is the optimal model parameter.

Our goal is to output a model $\mathbf{w}_{MN}$ minimizing the degree of \emph{forgetting} \cite{evron2022catastrophic} for $M$ tasks, i.e.
\begin{equation}\label{eq:forgetting}
    G(M) = \frac{1}{M} \sum_{m=1}^M \mathcal{L}_m (\mathbf{w}_{MN} ), \quad \text{where}
\end{equation}
\vspace{-5pt}
\begin{equation*}\label{eq:error}
    \mathcal{L}_m(\mathbf{w})=\frac{1}{2} \mathbb{E}_{(\mathbf{x}_m, y_m) \sim \mathcal{D}_m} \|\mathbf{x}_m^{\top} \mathbf{w}-y_m \|^2, \quad m \in \mathbb{M}
\end{equation*}

{$\mathbf{w}_{MN}$ represents the final output after sequentially training on $M$ tasks, each updated via SGD over $N$ iterations for each task.}
\cref{eq:forgetting} quantifies an average excess population risk on the final output $\mathbf{w}_{MN}$ across all tasks. For each task $m$, the loss $\mathcal{L}_m$ 
evaluate how well $\mathbf{w}_{MN}$ performs on it, thus assessing the degree of the model's forgetting on previous tasks in continual learning scenarios.

\begin{definition}[Data Covariance]
    Assume that each entry and the trace of the $\mathbb{E} [\mathbf{x}_m \mathbf{x}_{m}^{\top} ]$ are finite. Define $\mathbf{H}_m:=$ $\mathbb{E} [\mathbf{x}_m \mathbf{x}_{m}^{\top} ]$ as data covariance matrix.
\end{definition}

Let $\mathbf{H}_m$ denote the eigen decomposition of the data covariance for task $m$, given by $\mathbf{H}_m=\sum_i \lambda_m^i \mathbf{v}_m^i {\mathbf{v}_m^i}^{\top}$, where $(\lambda_m^i)_{i \geq 1}$ are eigenvalues in a nonincreasing order and $(\mathbf{v}_m^i)_{i \geq 1}$ are the corresponding eigenvectors. Define $\mathbf{H}_{m, k_1: k_2}$ as $\mathbf{H}_{m,k_1: k_2} := \sum_{k_1 < i \leq k_2} \lambda_m^i \mathbf{v}_m^i {\mathbf{v}_m^i}^{\top},$ and allow $k_2 = \infty$ to imply that $\mathbf{H}_{m,k: \infty} = \sum_{i > k}\lambda_m^i \mathbf{v}_m^i {\mathbf{v}_m^i}^{\top}$. 

\begin{definition}[Covariate Shift]\label{def:covariate shift}
    For each task $m$, the covariates $\mathbf{x}_{m,1}$, $\ldots$, $\mathbf{x}_{m,N}$ are i.i.d. drawn from $\mathcal{D}_m$.
\end{definition}

Compared to the concept of covariate shift in transfer learning \cite{pathak2022new}, \cref{def:covariate shift} provides a more general scenario applicable to a series of tasks $M\geq 2$. For simplicity, in our analysis, we assume that each task $m$ in our model consists of $N$ data points, differentiating it from transfer learning approaches that typically consider the total dataset size as $N$.

\begin{assumption}[Fourth moment conditions]\label{asm:fourth_moment}
   {Assume that for each task $m$, the expected fourth moment of covariates, denoted as $\mathcal{M}:=\mathbb{E}[\mathbf{x}_m \otimes \mathbf{x}_m \otimes \mathbf{x}_m \otimes \mathbf{x}_m]$, and 
    the expected covariance matrix $\mathbf{H}_m$ are finite.} Moreover:
    \begin{enumerate}[label=(\Alph*)]
    \item \label{asm:fourth_upper}There exists a constant $\alpha_m>0$ such that for any Positive Semi-Definite (PSD) matrix $\mathbf{A}$, the following holds: 
        $$
        \mathbb{E} [{\mathbf{x}_m} {\mathbf{x}_m}^{\top} \mathbf{A} {\mathbf{x}_m} {\mathbf{x}_m}^{\top} ] \preceq \alpha_m \cdot  \operatorname{tr}(\mathbf{H}_m \mathbf{A}) \mathbf{H}_m .
        $$
    \item \label{asm:fourth_lower} There exists a constant $\beta_m>0$, such that for every PSD matrix $\mathbf{A}$, the following holds:
      {\small  $$
        \mathbb{E} [{\mathbf{x}_m} {\mathbf{x}_m}^{\top} \mathbf{A} {\mathbf{x}_m} {\mathbf{x}_m}^{\top} ]-\mathbf{H}_m\mathbf{A}  \mathbf{H}_m\succeq \beta_m \cdot \operatorname{tr}(\mathbf{H}_m \mathbf{A}) \mathbf{H}_m .
        $$}
\end{enumerate}
\end{assumption} 

\begin{remark}\label{rem: asm}
    \cref{asm:fourth_moment} is a commonly employed assumption in the linear regression analysis utilizing SGD methods \cite{zou2021benign,wu2022last,wu2022power}, which is much weaker than the assumptions on the aforementioned related work. Specifically, it can be verified that \cref{asm:fourth_moment} holds with ${\alpha_m}=3$ and $\beta_m=1$ for Gaussian distribution discussed in \cite{asanuma2021statistical, lee2021continual, lin2023theory}. Additionally, \cref{asm:fourth_moment}\ref{asm:fourth_upper} can be relaxed to $\mathbb{E}\|\mathbf{x}_m\|_2^2 \leq \alpha_m \operatorname{tr}(\mathbf{H}_m)$ with $\mathbf{A}=\mathbf{I}$, where $\alpha_m \operatorname{tr}(\mathbf{H}_m)=1$ is assumed in \citealt{evron2022catastrophic}.
\end{remark}

\begin{assumption}[Well-specified noise]\label{asm:nosie}
    Assume that for each distribution of task $m$, the response (conditional on input covariates) is given by
    $y_m=\mathbf{x}_m^{\top} \mathbf{w}^*+z_m$, where $z_m \sim \mathcal{N} (0, \sigma^2 )$ and $z_m$ is independent with $\mathbf{x}_m$.
\end{assumption}

Similar to previous works, we assume that $z_m$ is some randomized noise that satisfies $\mathbb{E}[z_m|\mathbf{x}]=0$ and $\mathbb{E}[z_m^2]=\sigma^2$ for each task $m$.

\paragraph{Continual Learning via SGD}\label{par:sgd_min}
Suppose we train the model parameter $\mathbf{w}$ sequentially. Let $\mathbf{w}_{(m-1)N + N}$ represent the parameter state after the completion of training on task $m$, which also serves as the initial condition for the training of task $m+1$. Starting with $\mathbf{w}_0$ and employing a constant step size $\eta$, the model is updated by SGD for each task $m \in \mathbb{M}$ over $N$ iterations, with $t = 1, \ldots, N$:
{
\begin{equation}\label{eq:sgd_update}
    \begin{aligned}
    \mathbf{w}_{(m-1)N+t} &= \mathbf{w}_{(m-1)N+t-1}-\eta \cdot \mathbf{g}_{m,t}, \quad \text{and}\\
    \mathbf{g}_{m,t} &:= (\mathbf{x}_{m,t}^{\top}\mathbf{w}_{(m-1)N+t-1}-y_{m,t})\mathbf{x}_{m,t},
\end{aligned}
\end{equation}}
where $\mathbf{g}_{m, t}$ represents the gradient of the loss function at task $m$ and iteration $t$ for a given data point $ (\mathbf{x}_{m, t}, y_{m, t} )$.

Contrastingly, the minimum norm solution in linear regression, particularly relevant in overparameterized settings, aims to find a weight vector $\mathbf{w}$ that not only achieves zero training error but also possesses the minimal possible norm. Here, $\mathbf{w}_m$ represents the outcome post-training for task $m$, and it also serves as the starting point for training task $m+1$. The objective, beginning from an initial condition $\mathbf{w}_0 = \mathbf{0}$, is defined by the following optimization problem:
\begin{equation*}
\min _{\mathbf{w}} \|\mathbf{w}-\mathbf{w}_{m-1} \|_2, \quad \text {s.t. } (\mathbf{X}_m )^{\top} \mathbf{w}=\bm{y}_m,
\end{equation*}
where $\mathbf{X}_m := [\mathbf{x}_{m,1}, \ldots, \mathbf{x}_{m,N}] \in \mathbb{R}^{d \times N}$ and $\mathbf{y}_m = [y_{m,1}, \ldots, y_{m,N}] \in \mathbb{R}^{1 \times N}$. 
The update rules for each iteration follow as:
\begin{equation}
\mathbf{w}_m=\mathbf{w}_{m-1}+\mathbf{X}_m(\mathbf{X}_m^{\top} \mathbf{X}_m)^{-1}(\boldsymbol{y}_m-\mathbf{X}_m^{\top} \mathbf{w}_{m-1}),
\end{equation}
where highlights the computational intensity of inverting the matrix $(\mathbf{X}_m^{\top} \mathbf{X}_m)^{-1}$. This is particularly challenging for large datasets or overparameterized feature spaces. Unlike the minimum norm solution, SGD does not assume the existence of a unique, exact solution and is more adaptable to a variety of problems, including those with non-linear dynamics.
\section{Main Results}
Before presenting our upper bound, we shall establish the following notations to facilitate comprehension of the results.
{\small{
\begin{equation}\label{eq:def_notations}
    \left\{
    \begin{aligned}
        \Gamma_{(p,q)}^i &:= \prod_{j=p}^{q}  (1-\eta \lambda_j^i )^{2N}, \quad\bm{\Gamma}_{p}^{q} := \prod_{j=p}^q  (\mathbf{I}-\eta \mathbf{H}_j )^{2N}, \\
        \mathbf{U}_{k_m^*} &:= {\mathbf{I}_{m,{0: {k_m^*}}} +N\eta\mathbf{H}_{m,{{k_m^*}:\infty }}}, \quad \Lambda^i :=\sum_{m=1}^{M}\lambda_m^i,
    \end{aligned}
    \right.
\end{equation}
}}
\hspace{-0.1cm}where $(\lambda_m^i)_{i \geq 1}$ are eigenvalues of $\mathbf{H}_m$ in a nonincreasing order and $k_m^*=\max \{i: \lambda_m^i \geq \frac{1}{N \eta}\}$ represents the cut-off index for $\mathbf{H}_m$. {Here, $\Gamma_{(p, q)}^i$ and $\boldsymbol{\Gamma}_p^q$ can be regarded as a projection accumulation from task $p$ to task $q$, and basically capture the impact of the learning dynamic of previous tasks on the subsequent task. $\mathbf{U}_{k_m^*}$ is defined with respect to the cut-off index $k_m^*$ for each task's data covariance matrix $\mathbf{H}_m$ that captures both the dominant eigenvalues and the tail of the spectrum, and $\Lambda^i$ denotes the sum of the $i$-th eigenvalue across all tasks.}

In the following, we first provide our upper bound for the behavior of forgetting via SGD in the linear regression model.
\begin{theorem}[Upper Bound]\label{thm:main_up}
Consider a scenario where the model $\mathbf{w}$ undergoes training via SGD for $M$ distinct tasks, following a sequence $1, \ldots, M$. With a constant step size of $\eta \leq 1/R^2$ given that $R^2 = \max \{\alpha_m \operatorname{tr}(\mathbf{H}_m)\}_{m=1}^{M}$, each task $m$ is executed for $N$ iterations. Given that Assumptions \ref{asm:fourth_upper} and \ref{asm:nosie} are satisfied, the following will hold:
$$G(M) \leq \text{err}_{\text{var}} + \text{err}_{\text{bias}},$$
where the variance and bias errors are upper-bounded by 
\small
\begin{align*} 
    \text{err}_{\text{var}}  &\leq \frac{\sum_{m=1}^{M}}{M} \cdot \frac{\eta \sigma^2}{(1-\eta R^2)} \cdot D_1^{\text{eff}}, \\
    \text{err}_{\text{bias}} &\leq \frac{\sum_{k=1}^{M}}{M}   \|\mathbf{w}_0-\mathbf{w}^* \|_{\bm{\Gamma}_1^M\mathbf{H}_k}^2 \\
    &+   \frac{\sum_{m=1}^{M}}{M} \frac{ 2\alpha_m \eta^2 \cdot (D_{2}^{\text{eff}} + \Phi_1^{m-1}D_{3}^{\text{eff}} ) }{1-\eta \alpha_m \operatorname{tr}(\mathbf{H}_m)}\cdot  \|\mathbf{w}_0-\mathbf{w}^* \|_{\mathbf{U}_{k_m^*}}^2 \\
    & +   \frac{\sum_{m=1}^{M}}{M} \alpha_m \eta \cdot \|\mathbf{w}_0-\mathbf{w}^* \|_{\bm{\Gamma}_{1}^{M} \mathbf{H}_k (\mathbf{H}_m + {\Phi_1^{m-1}}\mathbf{I}) \cdot \mathbf{U}_{k_m^*}}^2,
\end{align*}
where the effective dimensions are given by 
\begin{equation}\label{eq:def_d_eff}
    \begin{aligned}
        &D_{1}^{\text{eff}}  :=  \sum_{i<k_m^*} \Gamma_{(m+1,M)}^i \Lambda^i+ N \eta \sum_{i>k_m^*} \Gamma_{(m+1,M)}^i  \lambda_m^i\Lambda^i \\ 
        &D_{2}^{\text{eff}}  :=\sum_{i<k_m^*} {\Gamma^i_{(1,M)}(\lambda_m^i)^2\Lambda^i}+N \eta \sum_{i>k_m^*} \Gamma^i_{(1,M)} (\lambda_m^i)^3\Lambda^i\\
        & D_{3}^{\text{eff}} :=  \sum_{i<k_m^*} {\Gamma_{(m,M)}^i (\lambda_m^i) \Lambda^i } +{\eta N} \sum_{i>k_m^*} \Gamma_{(m,M)}^i(\lambda_m^i)^2 \Lambda^i, 
    \end{aligned}
\end{equation}
with $k_m^*,\Gamma_{(p,q)}^i$ and $\bm{\Gamma}_{p}^{q}$ defined as in \Cref{eq:def_notations} and denoting $\Phi_1^{m-1} := \sum_{j=1}^{m-1} \prod_{k=1}^{j} \alpha_{k}\eta^j  \cdot \langle \mathbf{H}_{k-1}, \mathbf{I} - (\mathbf{I}- \eta \mathbf{H}_{m-1})^N \rangle \cdot  \langle \mathbf{H}_{j}, \mathbf{H}_{m}  \rangle$.
\end{theorem}

In \cref{thm:main_up}, we establish an upper bound on the forgetting behavior of a model trained using SGD in the continual learning with various data distribution settings. It highlights that the model's performance is influenced by both $\text{err}_{\text{var}}$ and $\text{err}_{\text{bias}}$, where $\text{err}_{\text{var}}$ stems from the inherent noise intrinsic to the model itself and $\text{err}_{\text{bias}}$ represents the bias associated with the initial value during the learning process. Notice that both of them are determined jointly by the spectrum of the covariance matrices as well as the stepsizes for continual learning. 

{To provide a more intuitive explanation, we explore a simplified scenario by setting $\eta=0$.
Specifically, this setting simplifies our analysis by reducing the error terms to only the first term in bias error, which appears to depend solely on the initial weight $\mathbf{w}_0$ and the data. However, this simplification might misleadingly imply that a minimal $\eta$ would result in optimal learning outcomes. A crucial aspect overlooked in this interpretation is the role of the projection term $\Gamma_1^M=\prod_{j=1}^M(\mathbf{I}-\eta \mathbf{H}_j)^{2 N}$, which becomes an identity matrix $\mathbf{I}$ when $\eta=0$. Thus, while setting $\eta=0$ eliminates other error terms, it also exacerbates the first term of bias error, potentially making it the most significant error contributor. Consequently, there exists a trade-off in choosing the step size.}

The subsequent theorem presents a nearly matching lower bound.
\begin{theorem}[Lower Bound] \label{thm:main_low}
Consider a scenario where the model $\mathbf{w}$ undergoes training via SGD for $M$ distinct tasks, following a sequence $1, \ldots, M$. With a constant step size of $\eta \leq 1/R^2$ given that $R^2 = \max \{\alpha_m \operatorname{tr}(\mathbf{H}_m)\}_{m=1}^{M}$, each task $m$ is executed for $N$ iterations. Given that Assumptions \ref{asm:fourth_lower} and \ref{asm:nosie} are satisfied, the following will hold:
$$G(M) \geq \text{err}_{\text{var}} + \text{err}_{\text{bia}},$$
where the variance and bias errors are lower bounded by 
{\small
\begin{align*} 
    &\text{err}_{\text{var}}  \geq \frac{\sum_{m=1}^{M}}{M} \cdot \frac{9 \eta^2 \sigma^2}{20} \cdot D_{1}^{\text{eff}},  \\
    &\text{err}_{\text{bias}} \geq \frac{\sum_{k=1}^{M}}{M}    \|\mathbf{w}_0-\mathbf{w}^* \|_{\bm{\Gamma}_1^M \mathbf{H}_k}^2 \\
    &+  \frac{\sum_{m=1}^{M}}{M} \cdot  \frac{\beta_m^2 \eta^2}{25} \cdot (D_{2}^{\text{eff}} + \hat{\Phi}_1^{m-1}D_{3}^{\text{eff}} ) \cdot  \|\mathbf{w}_0-\mathbf{w}^* \|_{ \mathbf{U}_{k_m^*}}^2 \\
    & + \frac{\sum_{m=1}^{M}}{M}\frac{\beta_m \eta^2}{5} \cdot  \|\mathbf{w}_0-\mathbf{w}^* \|_{(\mathbf{I}-\eta \mathbf{H}_m )^{2N} \bm{\Gamma}_{1}^{M} \mathbf{H}_k (\mathbf{H}_m + \hat{\Phi}_1^{m-1}\mathbf{I}) \cdot \mathbf{U}_{k_m^*}}^2
\end{align*}}
\hspace{-0.12cm}where the effective dimensions $k_m^*,\Gamma_{(p,q)}^i$ and $\bm{\Gamma}_{p}^{q}$ are the same as in \Cref{thm:main_up}, and {\small$\hat{\Phi}_1^{m-1}:=\sum_{j=1}^{m-1} \prod_{k=1}^{j} \beta_{k} (\frac{\eta}{2})^j  \cdot \langle\mathbf{H}_{k-1},( \mathbf{I}-(\mathbf{I}-\eta {\mathbf{H}_{m-1}})^{2N}) \rangle \cdot  \langle \mathbf{H}_{j}, \mathbf{H}_{m}  \rangle$}.
\end{theorem}

    {Analogous to the \cref{thm:main_up}, our lower bound also consists of the bias term and the variance term.
    It is noteworthy that our lower bound is tight with the upper bound in terms of variance term, differing only by absolute constants. Additionally, our lower bound closely matches the upper bound in terms of the bias term, with some differences arising from the following quantities
    \begin{equation*}
    \hat{\Phi}_1^{m-1}  \|\mathbf{w}_0-\mathbf{w}^* \|_{ \mathbf{U}_{k_m^*}}^2, \quad \|\mathbf{w}_0-\mathbf{w}^* \|_{(\mathbf{I}-\eta \mathbf{H}_m )^{2N}}.
    \end{equation*}
    Specifically, $\hat{\Phi}_1^{m-1}$ here differs from ${\Phi}_1^{m-1}$ in \cref{thm:main_up} only by a factor of constants (i.e. $\alpha_k$ and $\beta_k$ defined in \cref{asm:fourth_moment}). 
    The term {\small$\|\mathbf{w}_0-\mathbf{w}^* \|_{(\mathbf{I}-\eta \mathbf{H}_m )^{2N}}$} has a different subscript of $(\mathbf{I}-\eta \mathbf{H}_m )^{2N}$ compared to that of the upper bound. Nevertheless, it can be regarded as a part of the projection accumulation $\bm{\Gamma}_1^M$ that exists in the subscript of both results simultaneously.
    
    More importantly, we show that the upper and lower bounds converge, ignoring constant factors, under the conditions
    \begin{equation*}
        \|\mathbf{w}_0-\mathbf{w}^* \|_{ \mathbf{U}_{k_m^*}}^2 \lesssim \sigma^2, \quad \hat{\Phi}_1^{m-1} \lesssim O(1),
    \end{equation*}
    which can be satisfied that the signal-to-noise ratios $\|\mathbf{w}_0-\mathbf{w}^* \|_{ \mathbf{U}_{k_m^*}}^2/\sigma^2$ is bounded and the step size is appropriate small.
    }

    \section{Discussion}\label{sec:discuss}
    Building on \cref{thm:main_up} and \cref{thm:main_low}, we aim to offer a more comprehensive understanding of our findings from three key perspectives: 1) Technical Understanding Under Simplified Cases; 2) Comparison with Existing Work; 3) The Impact of Task Ordering and Parameters on Forgetting.
    
    \subsection{Technical Understanding Under Simplified Cases} \label{sec: technical}
    In this section, we demonstrate how to achieve a vanishing bound in the overparameterized regime. 
    
    Based on \cref{thm:main_up}, we consider a scenario where $\|\mathbf{w}_0-\mathbf{w}^*\|_2^2, \sigma^2 \lesssim 1$ and $\operatorname{tr}(\mathbf{H}_m) \simeq 1$ for each task $m$, implying a rapid decay in the spectrum of $\mathbf{H}_m$. To obtain a vanishing bound in the overparameterized regime, the effective dimension should hold that
    \begin{equation}\label{eq:d_eff_sim}
        \begin{aligned}
            D_{1}^{\text{eff}} \simeq D_{3}^{\text{eff}} &= o(\frac{MN}{e^{(M-m)}}), \\
            D_{2}^{\text{eff}} &= o(\frac{MN}{e^{M}}).
        \end{aligned}
    \end{equation}
    To meet the condition in \cref{eq:d_eff_sim}, for each task $\widetilde{m}$, let $k^{\dagger} = \min \{k_m^*, k_{\widetilde{m}}^*\}$ and $k^{\star} = \max \{k_m^*, k_{\widetilde{m}}^*$\}. It necessarily holds that 
    {\small
    \begin{equation}\label{eq:condition_d_eff}
        \begin{aligned} 
        \sum_{i<k^{\star}} \lambda_{\widetilde{m}}^i  &\simeq \sum_{i<k^{\star}} \lambda_m^i\lambda_{\widetilde{m}}^i\simeq \sum_{i<k^{\star}} (\lambda_m^i)^2 \lambda_{\widetilde{m}}^i= o({N}),\\
        \sum_{i>k^{\dagger}} \lambda_{\widetilde{m}}^i\lambda_m^i &\simeq \sum_{i>k^{\dagger}} (\lambda_m^i)^2\lambda_{\widetilde{m}}^i\simeq \sum_{i>k^{\dagger}} (\lambda_m^i)^3 \lambda_{\widetilde{m}}^i= o(\frac{1}{N}).
    \end{aligned}
    \end{equation}}
    To clarify \cref{eq:condition_d_eff}, let notice the crucial cut-off index $k^{\star}$ and $k^{\dagger}$, which divide the entire feature space into two $k^{\star}$-dimensional and $k^{\dagger}$-dimensional subspaces. For achieving a diminishing bound in overparameterized setting, it is necessary that the sum of eigenvalues for indices less than $k^{\star}$, denoted as $\sum_{i<k^{\star}}$, should be $o(N)$, and the sum of the tail eigenvalues for indices greater than $k^{\dagger}, \sum_{i>k^{\dagger}}$, should be $o(\frac{1}{N})$. These conditions are typically met when the dataset size $N$ is sufficiently large, or when a smaller step size $\eta$ is chosen dependent on $N$.
    Additionally, We note that the condition in \cref{eq:condition_d_eff} can be relaxed. In light of the definition of $k_m^*$, the eigenvalues for task $\widetilde{m}$ are truncated based on the following two scenarios: $\mathbf{1})$ $k_m^* \textless k_{\widetilde{m}}^*$ : Here, the cut-off for task $\widetilde{m}$ occurs earlier, resulting in an additional $(k_{\widetilde{m}}^*-k_m^*)$ dimensions of eigenvalues such that $\lambda_{\widetilde{m}}^i \geq 1 /(N \eta)$. To achieve a diminishing bound under this condition, it is necessary that $\sum_{k_m^* \leq i \leq k_{\tilde{m}}^*} \lambda_{\widetilde{m}}^i=$ $o(N )$. $\mathbf{2})$ $k_m^* \geq k_{\widetilde{m}}^*$ : In this case, the cut-off for task $\widetilde{m}$ occurs later, involving an additional $(k_m^*-k_{\widetilde{m}}^*)$ dimensions of eigenvalues where $\lambda_{\widetilde{m}}^i \leq 1 /(N \eta)$, achieving the same results. 
    
    In the under-parameterized regime, we even account for the worst-case scenario where $\lambda_m^i \geq \frac{1}{N \eta}$ for all index $i$ and task $m$, leading to a bound of $ D_{1}^{\text{eff}} \simeq D_{3}^{\text{eff}} = o(\frac{Md\lambda_m^1}{e^{(M-m)}}), D_{2}^{\text{eff}} = o(\frac{Md\lambda_m^1}{e^{M}}).$
    
    \subsection{Comparison with Existing work}\label{sec:compare}
    In this section, we will first explore the challenges and parallels between traditional/transfer learning and continual learning. Secondly, we examine how restrictive assumptions in previous studies might overshadow the impact of key factors, thereby affecting the overall understanding of forgetting in continual learning.
    
    Our results reveal that compared to traditional learning \cite{zou2021benign}, which typically involves a single task, and transfer learning \cite{wu2022power}, which usually incorporates two data distributions, the effective dimension in continual learning scenarios is more complex. Specifically, in our analysis, the term $\Lambda^i$ arises from a distinct measurement perspective (i.e. \emph{forgetting}), which requires us to consider how the final output aligns with all previously encountered tasks in the continual learning (i.e. $\mathbf{H}_m$ for all $m$). This is in contrast to both traditional training and transfer learning, where the evaluation metric is uniformly focused on performance against a single dataset (i.e. $\mathbf{H}_M$). Moreover, the multi-task nature of continual learning introduces unique challenges considering the bias iterates and variance iterates, where we refer to the proof in Appendix for more details.
     
    Given that our analysis, similar to theirs, characterizes bounds with the full eigenspectrum of the data covariance matrix, it follows that our derived results match their findings in several aspects: $\mathbf{1)}$ The cutoff index $k_m^*$ is uniquely determined for each task $m$ in continual learning, akin to the one in \citealt{zou2021benign,wu2022power}, where they identify corresponding indices $k_{\text {training }}^*$ and $k_{\text {test }}^*$. $\mathbf{2)}$ The projection terms ${\Gamma}_{(p,q)}^i$ and $\bm{\Gamma}_p^q$ also occur in transfer learning \cite{wu2022power}, showing how previous iterations/past learning is projected onto the future updates.

    Previous work \cite{evron2022catastrophic} also explored the dynamics of forgetting through the perspective of projection.
    We first revisit the findings presented by \citealt{evron2022catastrophic}. Considering a scenario where the number of iterations $N=1$, the update rule in their analysis can be reformulated as follows:
    \begin{equation}\label{eq:proj_evron}
        \mathbf{w}_{m} - \mathbf{w}^*= (\mathbf{I}-\eta_m \mathbf{x}_m\mathbf{x}_m^{\top})(\mathbf{w}_{m-1} - \mathbf{w}^*),
    \end{equation}
    where they incorporate the noiseless model assumption that $y_m = \mathbf{x}_m^{\top} \mathbf{w}^*$. As a result, the forgetting in \citealt{evron2022catastrophic} holds that
    {\small
    \begin{align*}
        &G(M)= \frac{1}{M} \sum_{m=1}^{M}\|\mathbf{x}_m(\mathbf{w}_M -\mathbf{w}^*)\|_2^2, \quad \text{given} \quad \|\mathbf{x}_m\|_2 \leq 1\\
        &\leq \frac{1}{M} \sum_{m=1}^{M}\|(\mathbf{I}-\eta_m\mathbf{x}_M \mathbf{x}_M^{\top})\ldots (\mathbf{I}-\eta_m\mathbf{x}_1 \mathbf{x}_1^{\top}) (\mathbf{w}_{0} - \mathbf{w}^*)\|_2^2,
    \end{align*}}
    indicating that the forgetting dynamic can be determined by the projection of $(\mathbf{I}-\eta_m\mathbf{x}_m \mathbf{x}_m^{\top})$, where $\eta_m=\|\mathbf{x}_m\|^{-2}$.
    However, compared to our analysis, their study exhibits several key differences in comparison to ours. $\mathbf{1)}$ The inherent model noise: \citealt{evron2022catastrophic} considers a noiseless model, where results in the absence of an additional iterative term $\mathbf{x}_m \cdot z_m$ related to noise in \cref{eq:proj_evron}. This omission leads to a lack of accumulative variance error in the evaluation of forgetting performance (i.e. $\text{err}_{\text{var}}$ in our analysis). It is noteworthy to mention that in numerous learning problems, the variance error often plays a dominant role in the total error \cite{jain2018parallelizing,zou2021benign,wu2022power}. $\mathbf{2)}$ The bounded norm $\|\mathbf{x}\|_2$: the assumption of the bounded norm, which omits the interaction with projection effects, is crucial in our analysis as the factor $\Lambda^i$ in \cref{thm:main_up} and \cref{thm:main_low}. $\mathbf{3)}$ Last iterate SGD results: \citealt{evron2022catastrophic} shows that, with a $\|\mathbf{x}_{m}\|_2^{-2}$ step size, their worst-case expected forgetting will become a dimension-dependent bound of $O(d/M)$. 
    This analysis, conducted under the overparameterized regime, suggests the occurrence of \emph{catastrophic forgetting}. In contrast, our results, as discussed earlier, offer a different perspective, suggesting the possibility of achieving a vanishing forgetting bound in overparameterized settings with certain conditions met. 

    It is noticed that \citealt{lin2023theory} also investigates the relationship between catastrophic forgetting and factors such as task sequence (order) and dimensionality. However, their results will tend to be vacuous in the under-parameterized setting since $(\mathbf{X}_m\mathbf{X}_m^{\top})^{-1}$, data matrix for task $m$, is non-invertible when employing minimum norm solution, as we discussed earlier in \cref{par:sgd_min}. Due to space constraints, a more extensive discussion will be provided in \cref{app:extension}.

    \begin{figure*}[t]
    \centering
    \begin{subfigure}{.24\textwidth}
        \centering
        \includegraphics[width=\linewidth]{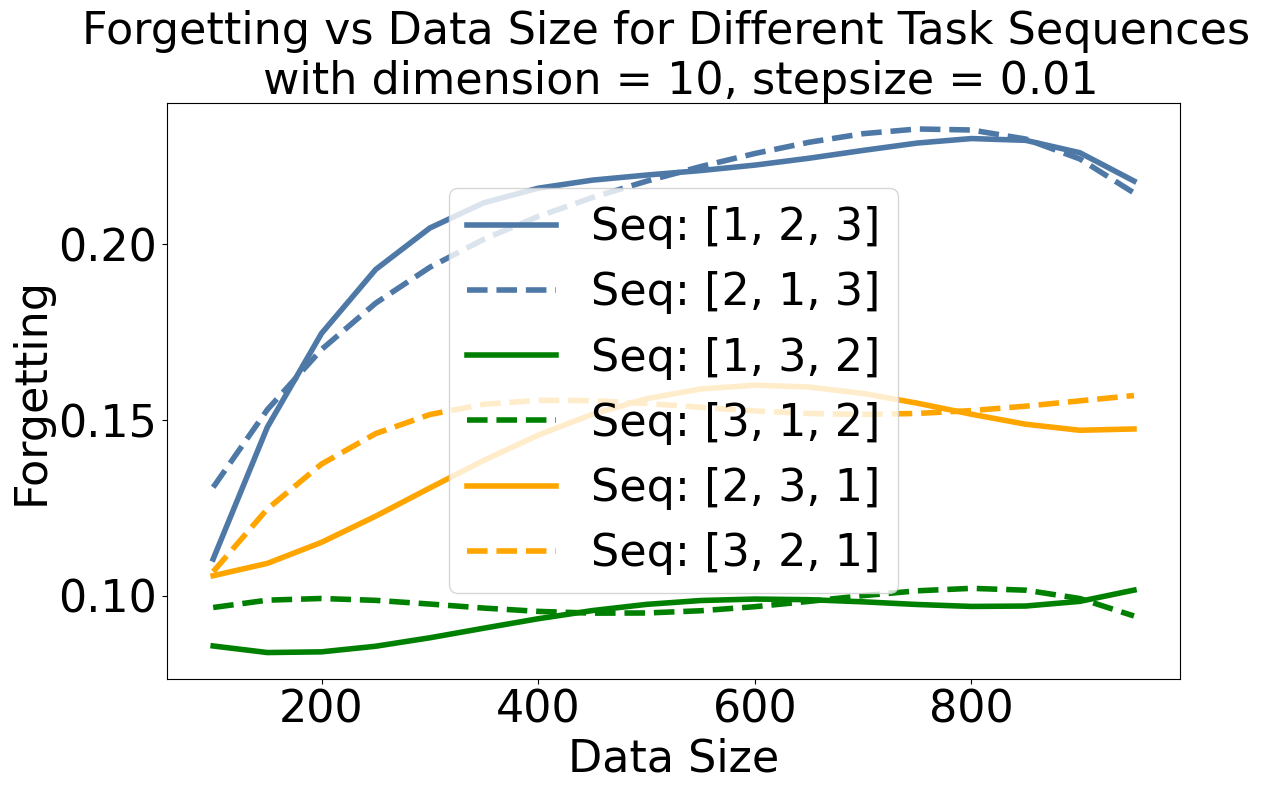}
        \caption{}
        \label{fig:low_a}
    \end{subfigure}%
    \begin{subfigure}{.24\textwidth}
        \centering
        \includegraphics[width=\linewidth]{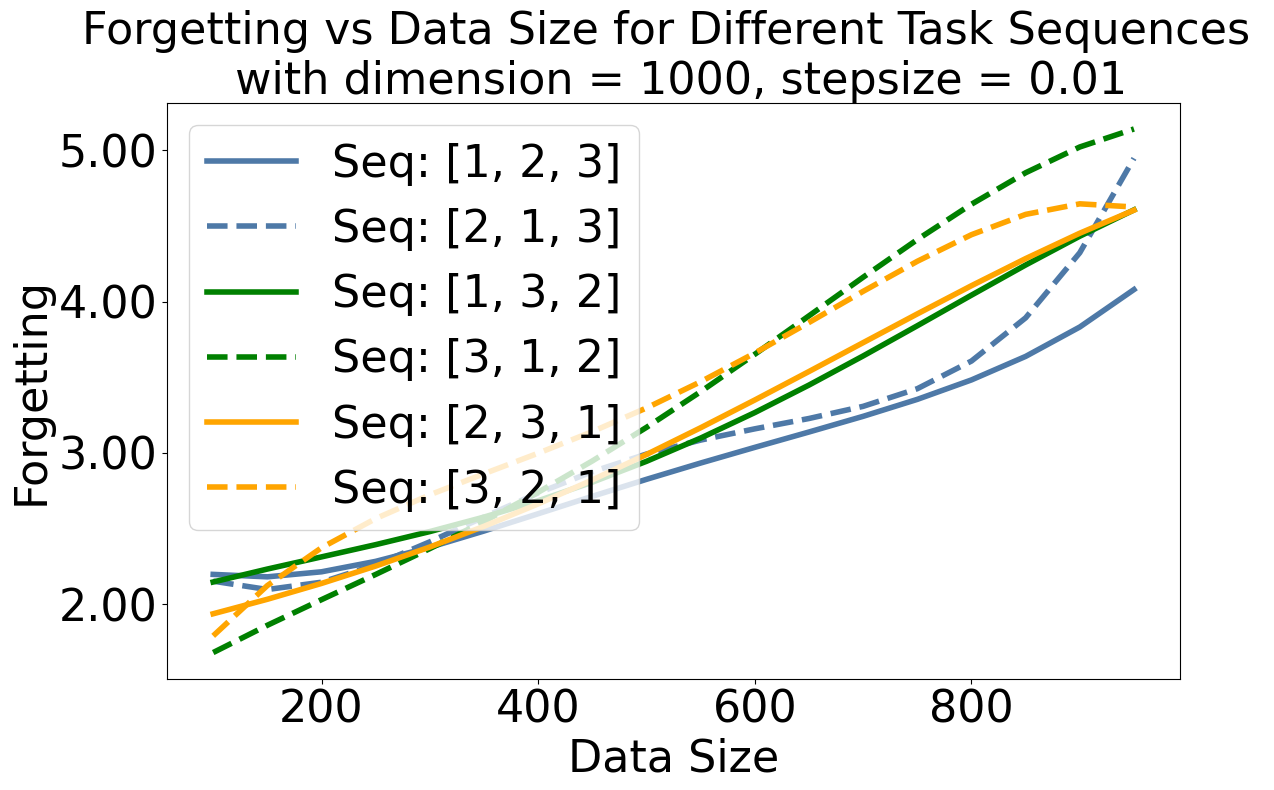}
        \caption{}
        \label{fig:high_b}
    \end{subfigure}
    \begin{subfigure}{.24\textwidth}
    \centering
    \includegraphics[width=\linewidth]{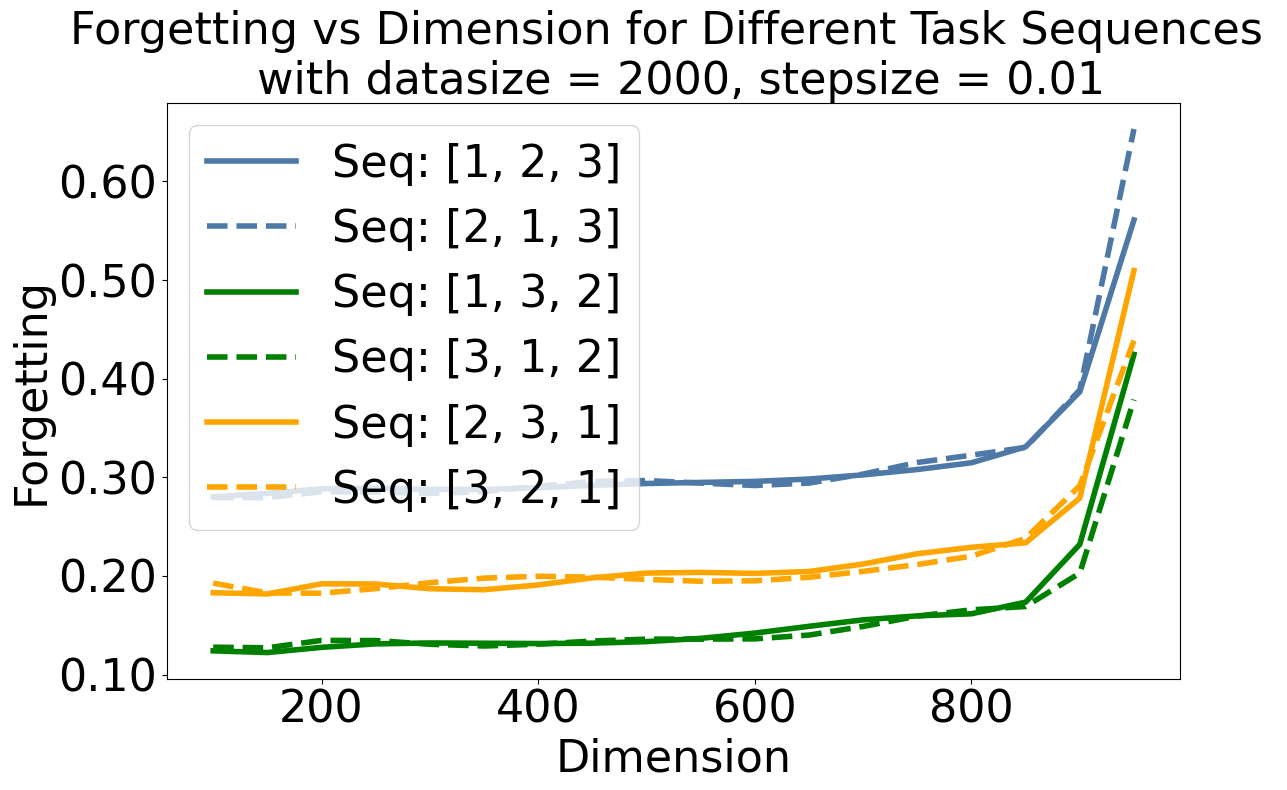}
    \caption{}
    \label{fig:high_d}
    \end{subfigure}%
    \begin{subfigure}{.24\textwidth}
        \centering
        \includegraphics[width=\linewidth]{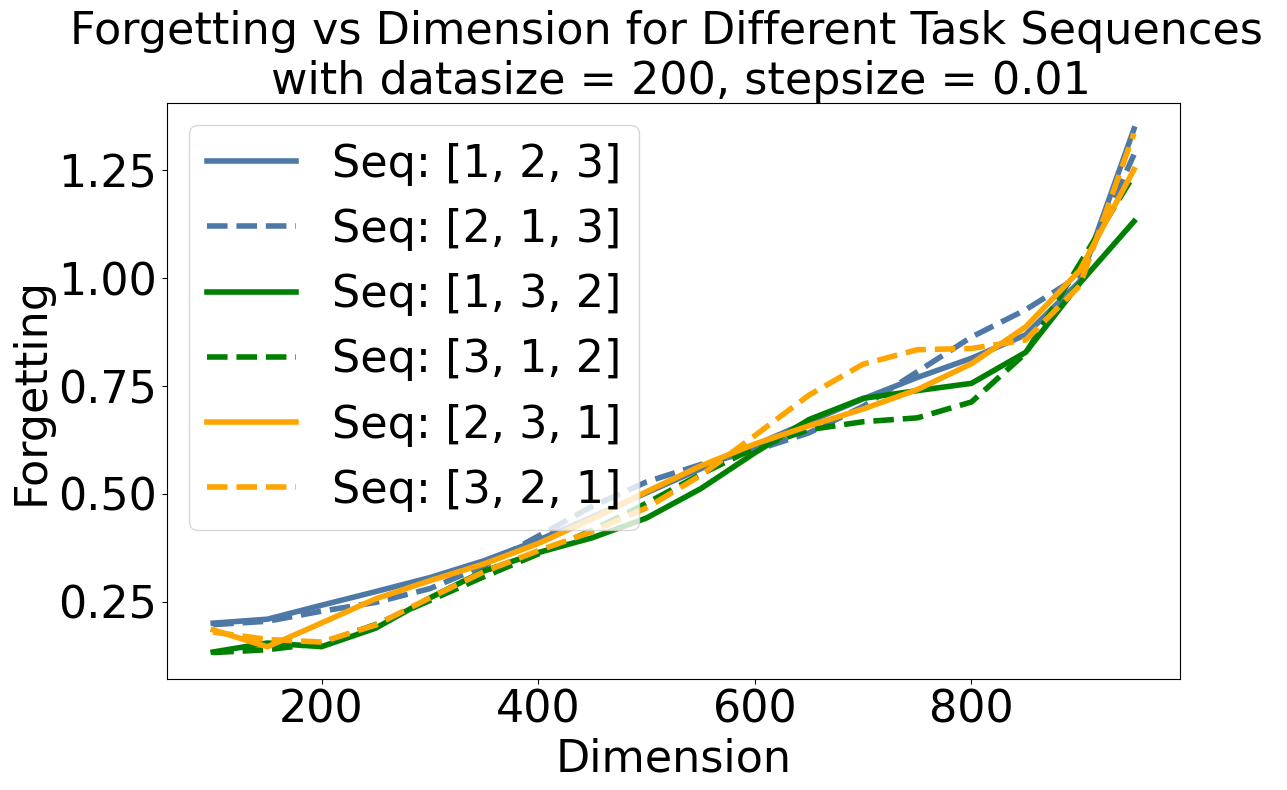}
        \caption{}
        \label{fig:dimension_small}
    \end{subfigure}
    \begin{subfigure}{.24\textwidth}
        \centering
        \includegraphics[width=\linewidth]{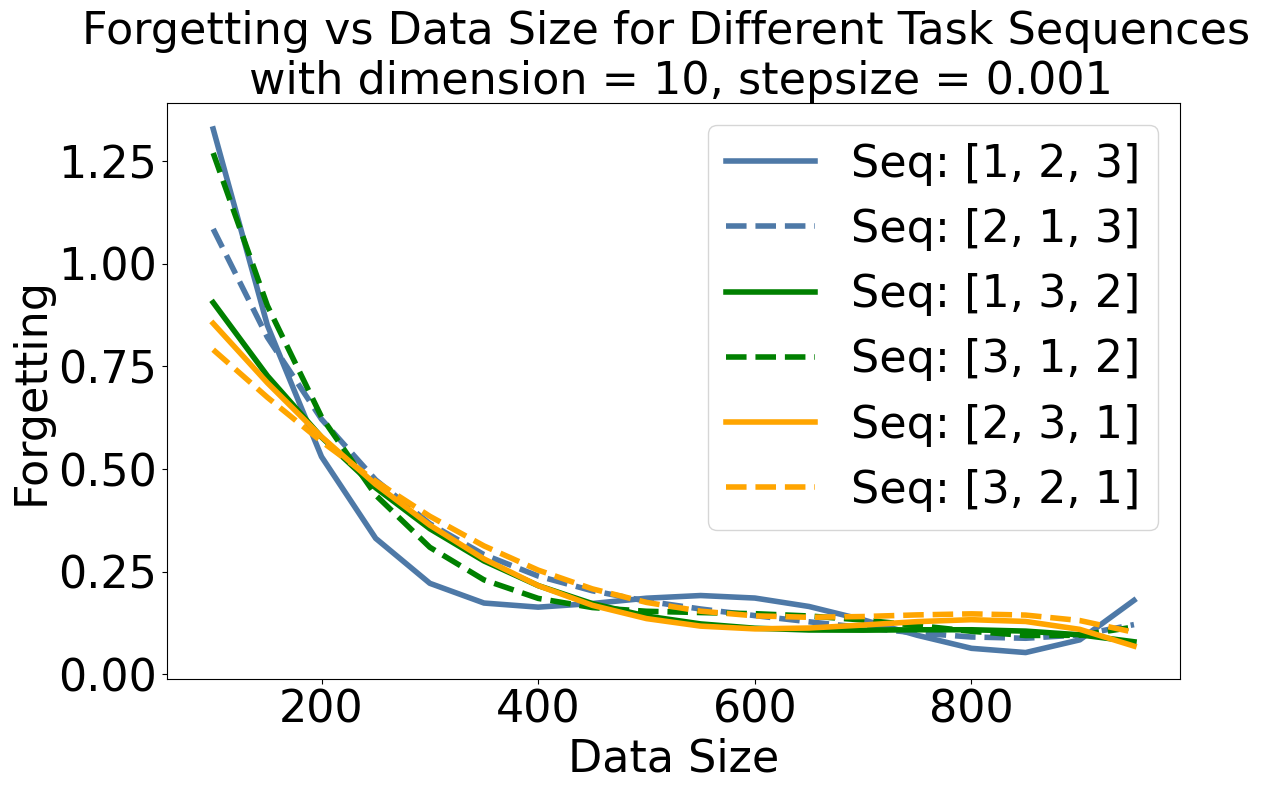}
        \caption{}
        \label{fig:step_low}
    \end{subfigure}
    \begin{subfigure}{.24\textwidth}
        \centering
        \includegraphics[width=\linewidth]{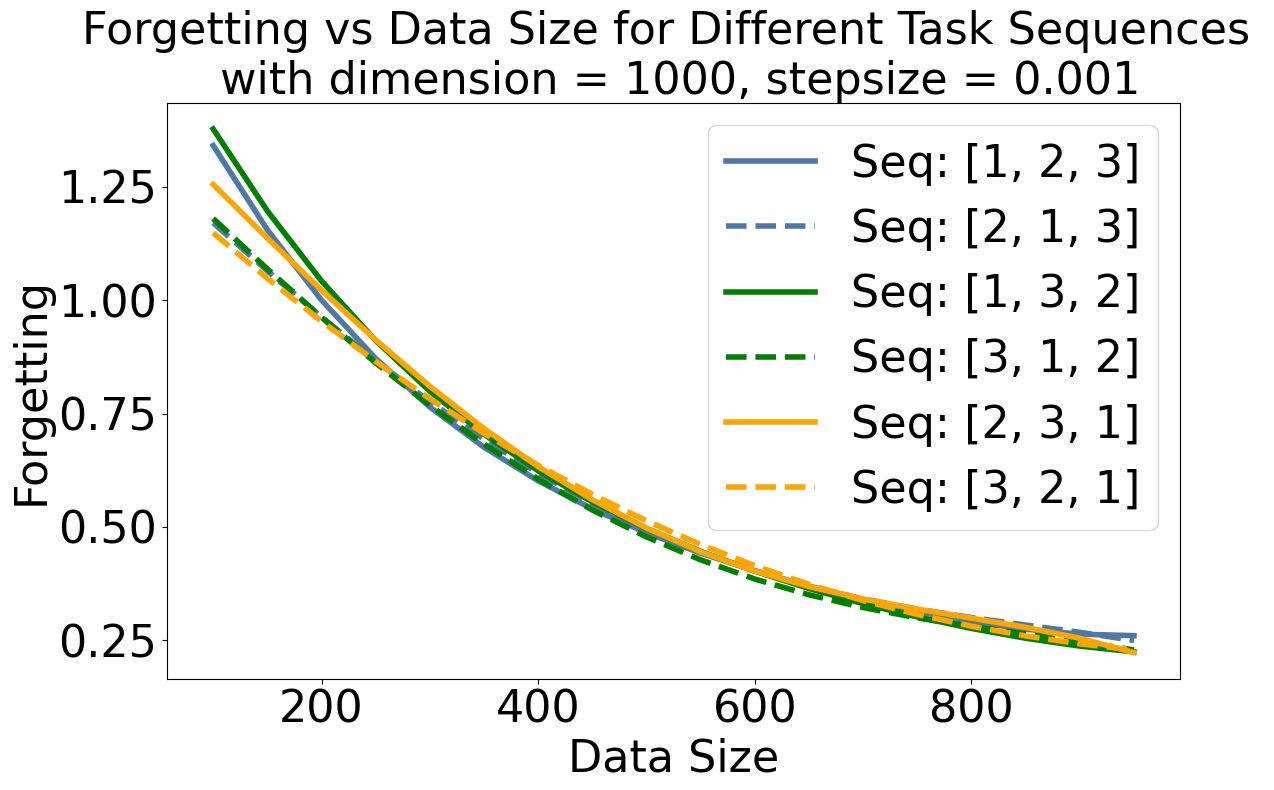}
        \caption{}
        \label{fig:step_high}
    \end{subfigure}
    \begin{subfigure}{.24\textwidth}
        \centering
        \includegraphics[width=\linewidth]{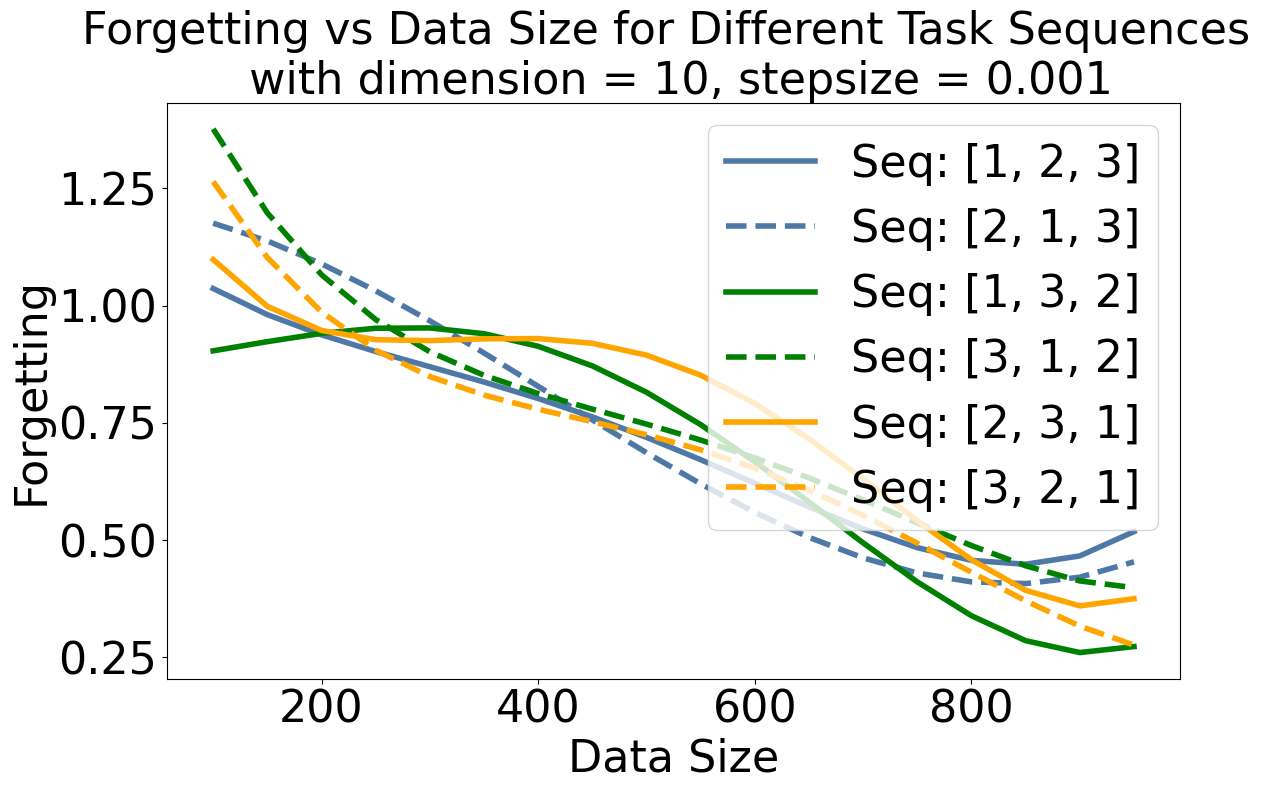}
        \caption{}
        \label{fig:dnn_step_low}
    \end{subfigure}
    \begin{subfigure}{.24\textwidth}
    \centering
    \includegraphics[width=\linewidth]{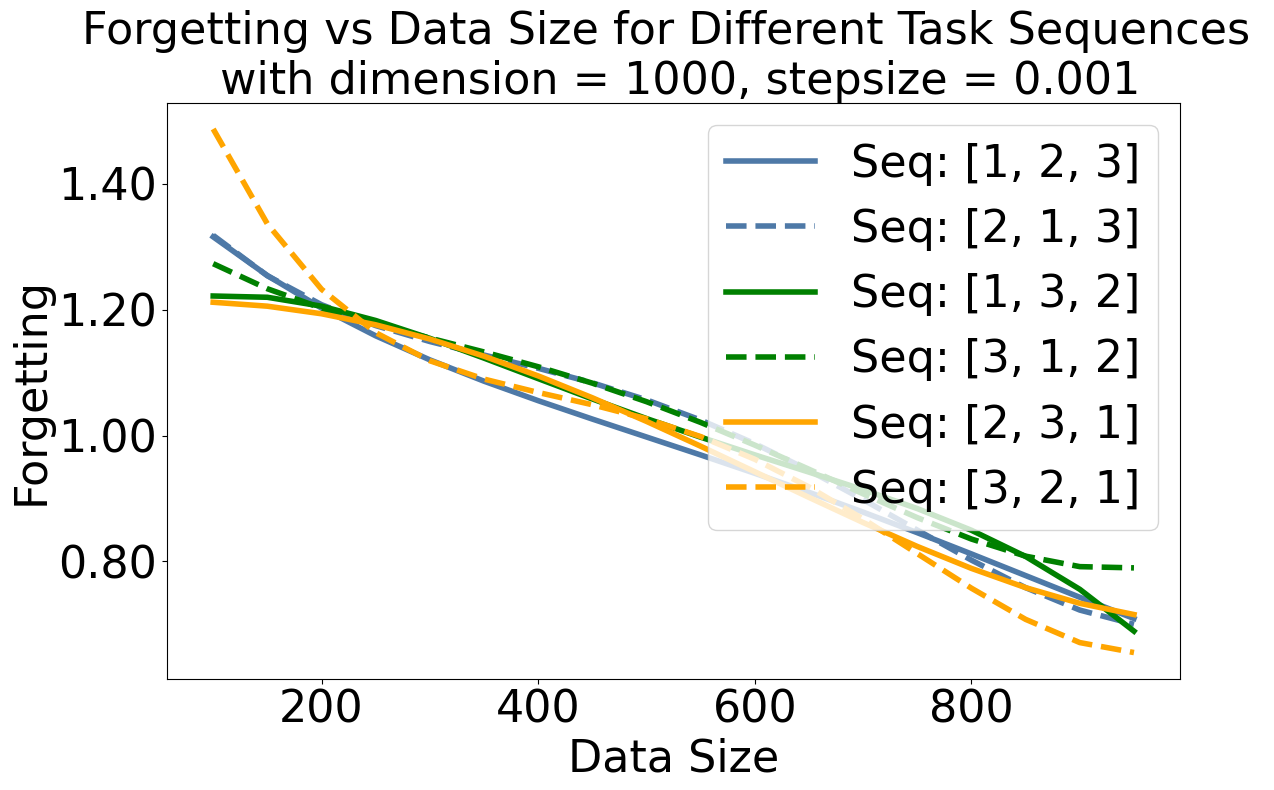}
    \caption{}
    \label{fig:dnn_step_high}
    \end{subfigure}

    \begin{subfigure}{.24\textwidth}
    \centering
    \includegraphics[width=\linewidth]{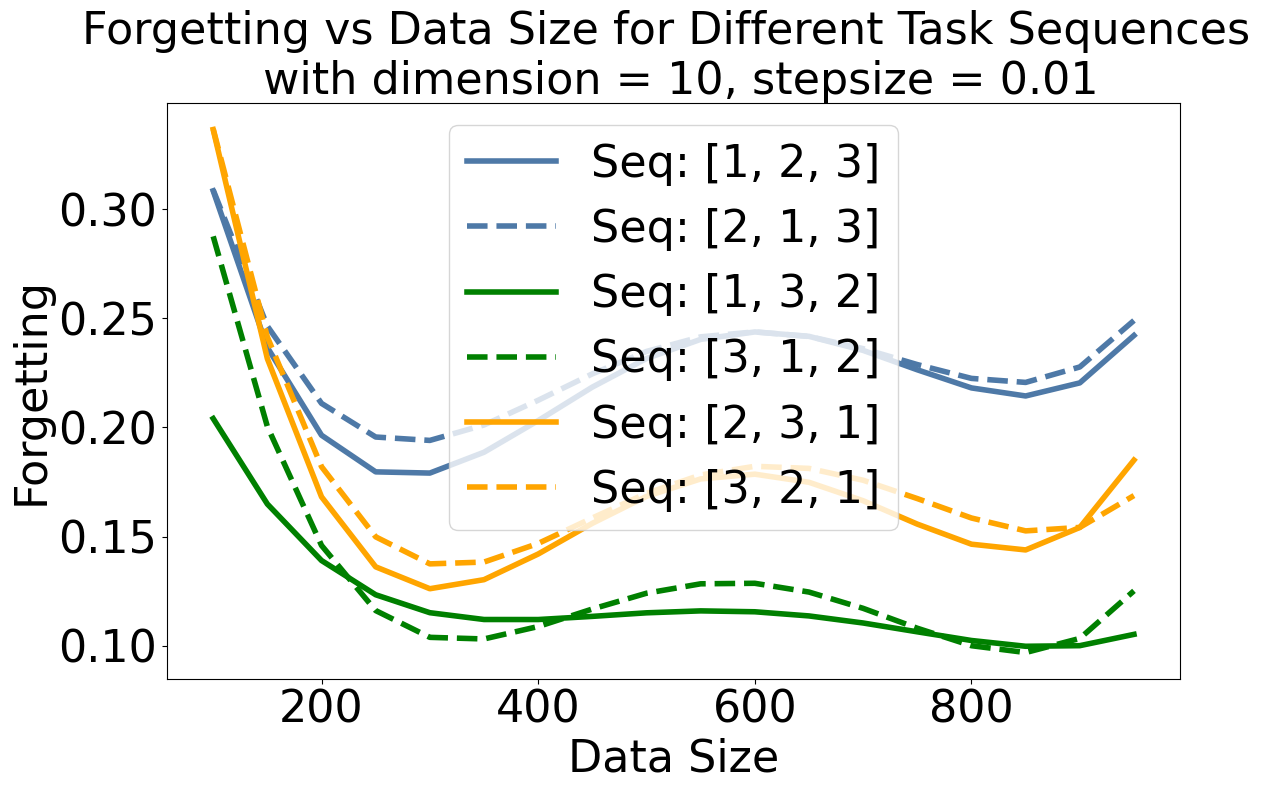}
    \caption{}
    \label{fig:dnn_low_a}
    \end{subfigure}%
    \begin{subfigure}{.24\textwidth}
        \centering
        \includegraphics[width=\linewidth]{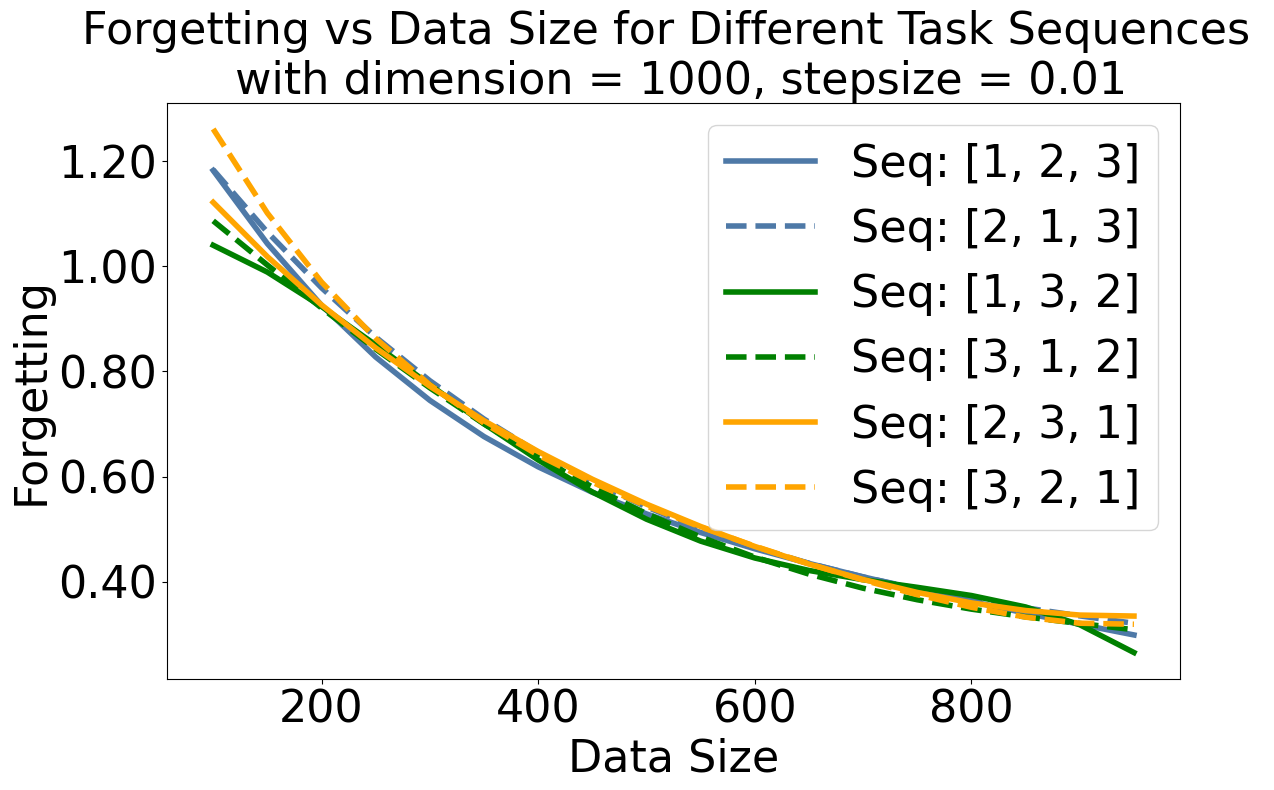}
        \caption{}
        \label{fig:dnn_high_b}
    \end{subfigure}
    \begin{subfigure}{.24\textwidth}
        \centering
        \includegraphics[width=\linewidth]{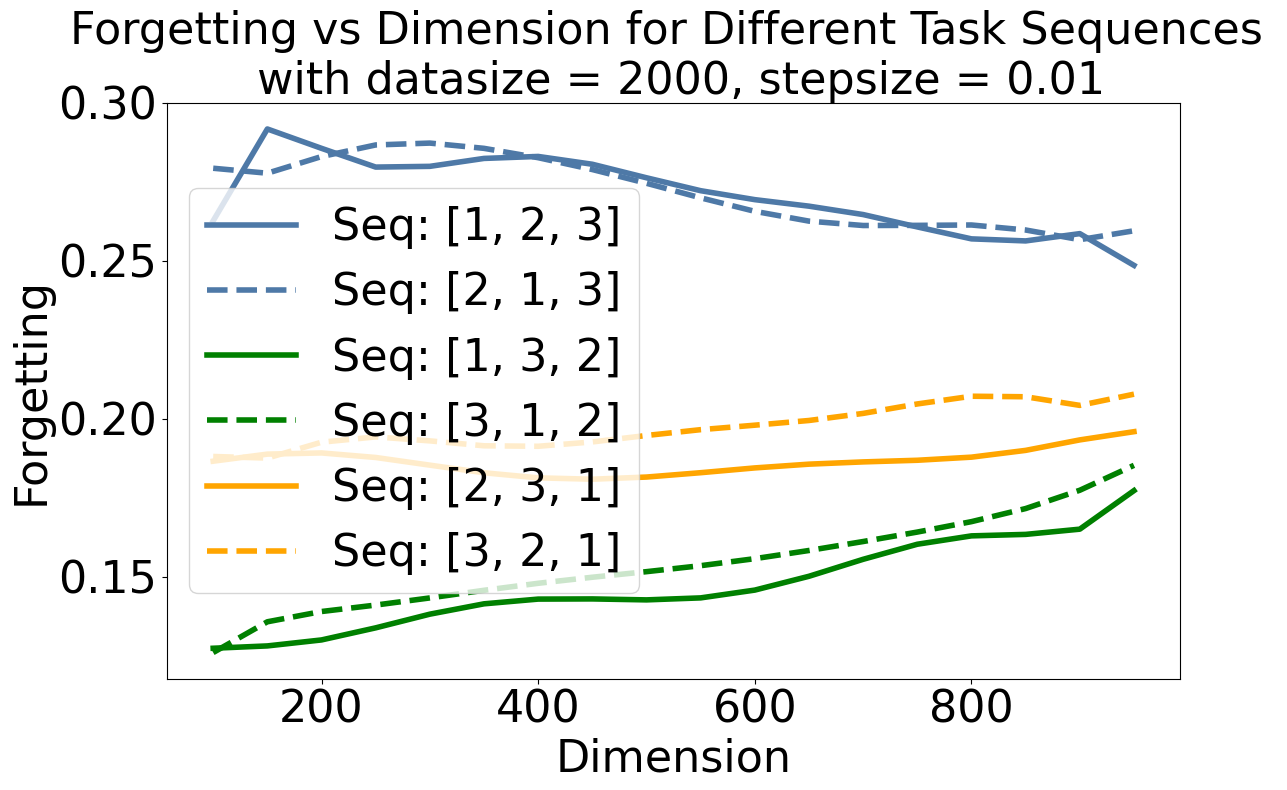}
        \caption{}
        \label{fig:dnn_high_d}
    \end{subfigure}%
    \begin{subfigure}{.24\textwidth}
        \centering
        \includegraphics[width=\linewidth]{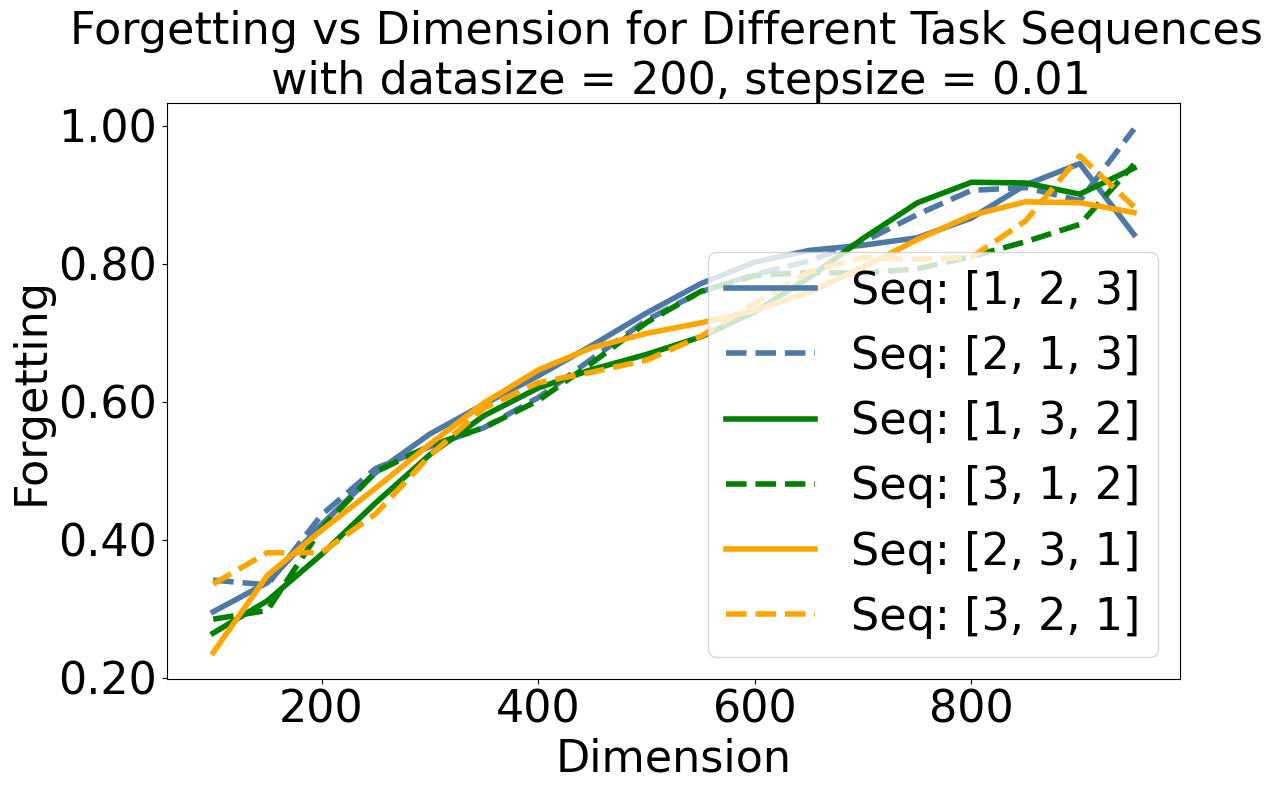}
        \caption{}
        \label{fig:dnn_dimension_small}
    \end{subfigure}
    
    \caption{Impact of Task Sequence and Algorithmic Parameters on Forgetting Behavior with Linear Regression Model and Deep Neural Networks. This figure presents the relationship between task sequence order and algorithmic parameters (data size, dimensionality, and step size) on the forgetting behavior observed in linear regression models (figures (a)-(f)) and deep neural networks (figures (g)-(l)). 
    Figures (a), (b), (i), and (j) illustrate how varying data sizes impact forgetting behavior for different task sequences, while figures (c), (d), (k), and (l) demonstrate the effect of changing dimensionality on forgetting. Lastly, Figures (e)-(h) demonstrate the influence of stepsize on the rate of forgetting across different model configurations.
    }
\end{figure*}

\subsection{The Impact of Task Ordering and Parameters on Forgetting}\label{sec:empirical_insights}
In the upcoming discussion, we will present theoretical insights derived from our results.

Notice that the bounds in \cref{thm:main_up} and \cref{thm:main_low} contain two crucial factors: the effective dimension $D^{\text{eff}}$ and the covariance accumulation $\hat{\Phi}_1^{m-1}$/${\Phi}_1^{m-1}$. We first discuss the effective dimension. Each $D^{\text{eff}}$ is consist of a projection term $\Gamma^i_{(m,M)}$ and the eigenvalues $\lambda_m^i$, with $\Lambda^i$ serving as the constant.
It can be observed that when data size $N$ approaches infinity, the projection term converges to $\frac{1}{e^{M-m}}$, implying that the eigenvalue will predominantly dictate the larger effective dimension with respect to $\frac{\lambda_m^i}{e^{M-m}}$. This observation highlights the substantial influence of eigenvalues on task sequence in continual learning. Specifically, it shows that \emph{\textbf{when data size is sufficiently large, task sequences organized in a way, where tasks associated with larger eigenvalues in their population data covariance matrix are trained later, exhibit more forgetting}}. 
Additionally, if the step size is appropriately small, the projection term stabilizes to a constant of less than 1, leading to similar outcomes as in the first scenario. It is noteworthy that these insights can not be derived from the existing work analysis due to their restrictive assumptions, such as Gaussian data distribution in \citealt{lee2021continual,asanuma2021statistical,lin2023theory} and minimum norm solution in \citealt{evron2022catastrophic,lin2023theory,swartworth2023nearly}.

The covariance accumulation term, $\hat{\Phi}_1^{m-1} / \Phi_1^{m-1}$, which includes the covariance matrices $\mathbf{H}_{j \leq m}$ and the step size $\eta$, plays a crucial role in demonstrating how previously acquired information is retained and influences the model's adaptability to new tasks. Notably, there is an interesting contradiction in the optimal accumulation order within $\hat{\Phi}_1^{m-1} / \Phi_1^{m-1}$ compared to the projection term in $\Gamma_{(m, M)}^i$. Specifically, earlier occurrence of $\mathbf{H}_j$ with larger expected eigenvalues tends to increase the degree of forgetting. Meanwhile, an important observation is that if the step size is sufficiently small, the impact of the covariance accumulation term becomes less significant. This interplay between the effective dimension and covariance accumulation elucidates the complexities inherent in continual learning scenarios.

\section{Empirical Stimulation}
In this section, we conduct experiments using synthetic data to validate our theoretical results and shed light on the intricate interplay between \emph{eigenvalues}, \emph{step size}, and \emph{dimensionality}.

\textbf{Experimental Setup}
In our study, we designed three distinct tasks, denoted as Tasks 1,2, and 3, each with a different feature space. During the initial simulations, the eigenvalues for the feature values of Tasks 1, 2, and 3 were set according to $\lambda_i=i^{-3}, \lambda_i=i^{-2}$, and $\lambda_i=i^{-1}$ respectively. To mimic real-world data imperfections, Gaussian noise with a standard deviation of 0.1 was added to the labels. We assessed the impact of task sequence on the model's tendency to forget by evaluating six different task orders: [1,2,3], [2,1,3], [1, 3, 2], [3, 1, 2], [2, 3, 1], and [3, 2, 1].

\subsection{Linear Regression}\label{sec:exp_linear}
\textbf{Training and Evaluation}
For this experiment, a linear regression model was trained using Stochastic Gradient Descent (SGD) with a learning rate of 0.01 or 0.001. The model was tested in both low-dimensional (10 input features) and high-dimensional (1000 input features) settings. Each task sequence underwent training with various data sizes, ranging from 100 to 950 in increments of 50, and each task was trained for five epochs. The performance of the model was evaluated on each task to calculate the average excess risk (\cref{eq:forgetting}), quantifying the degree of forgetting the model experienced.

\textbf{Impact of Eigenvalue Sequencing}
The observations from \cref{fig:low_a} and \cref{fig:high_d} reveal the significant impact of eigenvalue sequencing on forgetting behavior in the underparameterzied regime. Notably, task sequences that are arranged such that tasks with larger eigenvalues (i.e. Task 3 in our case, characterized by $\lambda_i=i^{-1}$ ) are trained later in the learning process tend to result in increased forgetting. This empirical finding aligns well with our theoretical analysis (the term $\frac{\lambda_m^i}{e^{M-m}}$ discussed in \cref{sec:empirical_insights}). In an under-parameterized setting, or when the eigenvalues decay rapidly, the effective dimension — crucial in determining the model's forgetting performance - is largely influenced by the eigenvalues. Such a pattern is intuitive as when tasks with larger eigenvalues are trained later, the model might overfit these tasks due to their high variance.

\textbf{Impact of Dimensionality}
Our results, depicted in \cref{fig:high_d} and \cref{fig:dimension_small}, show that in under-parameterized scenarios, performance remains relatively unaffected by an increase in dimensionality. However, in over-parameterized settings, the model tends to exhibit increased forgetting as dimensionality rises, particularly when the data size is kept constant. This highlights the varying impact of dimensionality on model performance in different parameterization contexts. In higher-dimensional settings, the influence of the projection term $\Gamma_{(p, q)}^i$, as shown in \cref{thm:main_up}, diminishes in comparison to the impact of $\Lambda^i$ and $\lambda_i$. Consequently, as the number of features in the model increases, the sequence in which tasks are presented becomes less significant in determining the model's forgetting behavior. This shift implies that, in high-dimensional scenarios, the inherent complexity and the distribution of eigenvalues of the feature space play a more critical role than the sequence of tasks, influencing the model's learning and retention capabilities. 

\textbf{Impact of Step-size}
Our results, depicted in \cref{fig:step_low} and \cref{fig:step_high}, reveal that a smaller step size effectively reduces forgetting in various task sequences and across different dimensionalities. This trend is especially noticeable in high-dimensional feature spaces, where a reduced step size markedly lowers the rate of forgetting.
This observation is in line with the theoretical insights provided in \cref{thm:main_up} and \cref{thm:main_low}, as smaller step sizes may lead to more refined updates during training, allowing the model to incrementally adjust to new tasks while preserving knowledge from previous ones. 

\subsection{Implication on DNNs}

Intriguingly, our next discussion will adopt the same data generation and task setup as outlined in \cref{sec:exp_linear}, but shift our focus to a different Neural Network model. This model comprises an input layer, a hidden layer with ten neurons, and an output layer, and it undergoes a training process akin to that of linear regression. 

\textbf{Impact of Eigenvalue Sequencing}
In our studies with Deep Neural Networks (DNNs), we still find that task sequences, ending with tasks having larger eigenvalues, tend to exhibit increased forgetting, especially in under-parameterized settings, similar to linear regression models. This indicates that the tendency of overfitting observed in linear models, particularly when tasks with larger eigenvalues are trained later in the sequence, may occur in DNNs as well.

\textbf{Impact of Dimensionality}
Our results also reveal the consistent behaviors between DNNs and linear regression concerning dimensionality. In under-parameterized scenarios (\cref{fig:dnn_high_d}), forgetting remains stable despite increased dimensionality, while in over-parameterized settings (\cref{fig:dnn_dimension_small}), higher dimensionality leads to more forgetting when data size is fixed. However, the adverse effects of higher dimensions can be alleviated by expanding the dataset size, as demonstrated in \cref{fig:dnn_high_b}. It is a notable contrast to linear regression, which suggests that the complex structures of DNNs are better suited to manage and learn from high-dimensional data in continual learning scenarios. {The different behaviors observed between DNNs
and linear regression models will be a potentially interesting direction for future work}.

\textbf{Impact of Step Size}
Our results, depicted in \cref{fig:step_low} and \cref{fig:step_high}, indicate that in under-parameterized settings, a smaller step size significantly lessens the influence of task sequences on forgetting, while in models with high-dimensional features, forgetting can be mitigated even without adjusting the step size. 
\vspace{-2.5pt}
\section{Conclusion}
{In this work, we contribute to the understanding of catastrophic forgetting in continual learning via a multi-step SGD algorithm. Our theoretical analysis establishes bounds that illustrate the impact of various factors on forgetting such as data covariance matrix spectrum, step size, data size, and dimensionality, which can not be fully captured in previous studies due to their restrictive assumptions. This theoretical understanding is further substantiated through simulations conducted in linear regression models and Deep Neural Networks, which corroborate our theoretical insights.}
\section*{Impact Statements}
This paper presents work whose goal is to advance the field of Machine Learning. There are many potential societal consequences of our work, none of which we feel must be specifically highlighted here.

\section*{Acknowledgments}
The research of Meng Ding and Jinhui Xu was supported in part by KAUST through grant CRG10-4663.2. Di Wang was supported in part by the baseline funding BAS/1/1689-01-01, funding from the CRG grand URF/1/4663-01-01, REI/1/5232-01-01,  REI/1/5332-01-01,  FCC/1/1976-49-01 from CBRC of King Abdullah University of Science and Technology (KAUST). Di Wang was also supported by the funding RGC/3/4816-09-01 of the SDAIA-KAUST Center of Excellence in Data Science and Artificial Intelligence (SDAIA-KAUST AI).

\bibliography{example_paper}
\bibliographystyle{icml2024}

\newpage
\appendix
\onecolumn
\section{Support Lemmas}
\textbf{Notations}

For two matrices $\mathbf{A}$ and $\mathbf{B}$, their inner product is defined as $\langle\mathbf{A}, \mathbf{B}\rangle:=\operatorname{tr} (\mathbf{A}^{\top} \mathbf{B} )$. For each task $m \in  [M ]$, we define the following linear operators:
$$
\begin{gathered}
\mathcal{I}=\mathbf{I} \otimes \mathbf{I}, \quad \mathcal{M}_m=\mathbb{E}[\mathbf{x}_m \otimes \mathbf{x}_m \otimes \mathbf{x}_m \otimes \mathbf{x}_m], \quad \widetilde{\mathcal{M}}_m=\mathbf{H}_m \otimes \mathbf{H}_m, \\
\mathcal{T}=\mathbf{H} \otimes \mathbf{I}+\mathbf{I} \otimes \mathbf{H}_m-\eta \mathcal{M}_m, \quad \widetilde{\mathcal{T}}=\mathbf{H}_m \otimes \mathbf{I}+\mathbf{I} \otimes \mathbf{H}_m-\eta \mathbf{H}_m \otimes \mathbf{H}_m .
\end{gathered}
$$

We use the notation $\mathcal{O} \circ \mathbf{A}$ to denote the operator $\mathcal{O}$ acting on a symmetric matrix $\mathbf{A}$. For example, with these definitions, we have that for a symmetric matrix $\mathbf{A}$,
$$
\begin{gathered}
\mathcal{I} \circ \mathbf{A}=\mathbf{A}, \quad \mathcal{M}_m \circ \mathbf{A}=\mathbb{E} [ ({\mathbf{x}_m}^{\top} \mathbf{A} {\mathbf{x}_m} ) {\mathbf{x}_m} {\mathbf{x}_m}^{\top} ], \quad \widetilde{\mathcal{M}}_m \circ \mathbf{A}=\mathbf{H}_m \mathbf{A} \mathbf{H}_m \\
(\mathcal{I}-\eta \widetilde{\mathcal { T }}_m) \circ \mathbf{A}=(\mathbf{I}-\eta \mathbf{H}_m) \mathbf{A}(\mathbf{I}-\eta \mathbf{H}_m)\\
(\mathcal{I}-\eta \mathcal{T}_m) \circ \mathbf{A}=\mathbb{E} [ (\mathbf{I}-\eta {\mathbf{x}_m} {\mathbf{x}_m}^{\top} ) \mathbf{A} (\mathbf{I}-\eta {\mathbf{x}_m} {\mathbf{x}_m}^{\top} ) ]
\end{gathered}
$$

It can be readily understood that the following properties are satisfied:

\begin{lemma}[\cite{zou2021benign}] \label{lem:properties}
    An operator $\mathcal{O}$, when defined on symmetric matrices, is termed a Positive Semi-Definite (PSD) mapping if $\mathbf{A} \succeq 0$ implies $\mathcal{O} \circ \mathbf{A} \succeq 0$. Consequently, for each task $m \in  [ M ]$ we have:
    \begin{itemize}
        \item [1.] $\mathcal{M}_m$ and $\widetilde{\mathcal{M}}_m$ are both PSD mappings.
        \item [2.] $\mathcal{M}_m-\widetilde{\mathcal{M}}_m$ and $\widetilde{\mathcal{T}}_m-\mathcal{T}_m$ are both PSD mappings.
        \item [3.] $\mathcal{I}-\eta\mathcal{T}_m$ and $\mathcal{I}-\eta\widetilde{\mathcal{T}}_m$ are both PSD mappings.
        \item [4.] If $0<\eta<1 / \lambda_m^1$, then $\widetilde{\mathcal{T}}^{-1}$ exists, and is a PSD mapping.
        \item [5.] If $0<\eta<1 /(\alpha_m \operatorname{tr}(\mathbf{H}_m))$, then $\mathcal{T}_m^{-1} \circ \mathbf{A}$ exists for PSD matrix $\mathbf{A}$, and $\mathcal{T}_m^{-1}$ is a PSD mapping.
    \end{itemize}
\end{lemma}

Then for the SGD iterates, we can consider their associated bias iterates and variance iterates:
\begin{align} \label{eq:decom_iter}
& \begin{cases}\mathbf{B}_0= (\mathbf{w}_0-\mathbf{w}^* ) (\mathbf{w}_0-\mathbf{w}^* )^{\top} , \\ 
\mathbf{B}_{(m-1)N+t+1}= (\mathcal{I}-\eta \mathcal{T}_{m} (\eta ) ) \circ \mathbf{B}_{(m-1)N+t}; \end{cases} \\
& \begin{cases}\mathbf{C}_0=\mathbf{0}, & \\ 
\mathbf{C}_{(m-1)N+t+1}= (\mathcal{I}-\eta \mathcal{T}_{\mathbf{H}_m} (\eta  ) ) \circ \mathbf{C}_{(m-1)N+t}+\eta^2 \bm{\Sigma}_{\mathbf{H}_m};
\end{cases} 
\end{align}
where  $t=0, \ldots, N-1$ and $m=1, \ldots, M$. 

\begin{lemma}[Bias-variance decomposition]\label{lem:risk_decom}
    Suppose that Assumption \ref{asm:nosie} holds. Then we have:
    $$
    \mathbb{E} [\operatorname{ExcessRisk} (\mathbf{w}_{MN} ) ]=\frac{1}{2} \langle\mathbf{H}, \mathbf{B}_{MN} \rangle+\frac{1}{2} \langle\mathbf{H}, \mathbf{C}_{MN} \rangle .
    $$
\end{lemma}

\section{Variance Error}
\subsection{Upper Bound}
The assumption presented below can be inferred from \ref{asm:fourth_moment} by setting $\mathbf{A} = \mathbf{I}$, given that $R^2 = \max \{\alpha_m \operatorname{tr}(\mathbf{H}_m)\}_{m=1}^{M}$.

\begin{assumption}[Relaxed version]
    For each task $m$, there exists a constant $R \geq 0$ such that:
$$
\mathbb{E}_{\mathbf{x} \sim \mathcal{D}_{m}} [\mathbf{x} \mathbf{x}^{\top} \mathbf{x} \mathbf{x}^{\top} ] \preceq R^2 \mathbf{H}_m.
$$
\end{assumption}

\begin{lemma}\label{lem:c_t_bound}
    Suppose Assumptions \ref{asm:fourth_moment} and \ref{asm:nosie} hold with step size $\eta \leq 1/R^2$, then it holds that:
    $$
    \mathbf{C}_t \leq \frac{\eta \sigma^2}{1-\eta R^2} \mathbf{I}, \quad \text { for every } t=0,1, \ldots, MN
    $$
\end{lemma}

\begin{proof}[{\bf Proof}]
    This lemma is derived directly from the Lemmas in \cite{jain2018parallelizing, zou2021benign}. To ensure completeness, we include a proof as follows.
    
    We prove the lemma via induction. Initially, for $t = 0$, it is evident that $\mathbf{C}_0 = \mathbf{0} \preceq \frac{\eta \sigma^2}{1 - \eta R^2} \mathbf{I}$. Now, assuming that $\mathbf{C}_{t} \preceq \frac{\eta \sigma^2}{1 - \eta R^2} \mathbf{I}$, let us examine $\mathbf{C}_{t+1}$ in light of \cref{eq:decom_iter}. When $0 \leq t \leq N-1$, for each task $m$, it implies:
    \begin{align} \notag
        \mathbf{C}_{(m-1)N+t+1} & = (\mathcal{I}-\eta \mathcal{T}_{\mathbf{H}_m} (\eta  ) ) \circ \mathbf{C}_{(m-1)N+t}+\eta^2 \bm{\Sigma}_{\mathbf{H}_m} \\ 
        & \preceq \frac{\eta \sigma^2}{1 - \eta R^2} \mathbf{I} \cdot  (\mathcal{I}-\eta \mathcal{T}_{\mathbf{H}_m} (\eta  ) ) \circ \mathbf{I} + \eta^2\sigma^2 \mathbf{H}_m \\\notag
        & \preceq  \frac{\eta \sigma^2}{1 - \eta R^2} \cdot  ( \mathbf{I} - 2\eta \mathbf{H}_m + \eta^2 R^2 \mathbf{H}_m  )  + \eta^2\sigma^2 \mathbf{H}_m \\ \notag
        & \preceq  \frac{\eta \sigma^2}{1 - \eta R^2} \cdot \mathbf{I}.
    \end{align}
\end{proof}

\begin{lemma}
    Suppose Assumptions \ref{asm:fourth_moment} and \ref{asm:nosie} hold with step size $\eta \leq 1/R^2$, then it holds that:
    \begin{align}\notag
        \mathbf{C}_{MN} & \preceq \prod_{m=2}^{M}  (\mathcal{I}-\eta \widetilde{\mathcal{T}}_{\mathbf{H}_m} (\eta  ) )^N \circ \mathbf{C}_{N} + \sum_{m=1}^{M-1}\prod_{j=m+1}^M  (\mathcal{I}-\eta \widetilde{\mathcal{T}}_{\mathbf{H}_j} (\eta  ) )^N \mathbf{P}_m + \mathbf{P}_M,
    \end{align}
    where $\mathbf{P}_m =\frac{\eta^2 \sigma^2}{1-\eta R^2} \sum_{t=0}^{N-1}  (\mathcal{I}-\eta \widetilde{\mathcal{T}}_{\mathbf{H}_m}(\eta) )^t \circ \mathbf{H}_m$ and $\mathbf{C}_N \preceq \frac{\eta \sigma^2}{1-\eta R^2} \cdot (\mathbf{I}-(\mathbf{I}-\eta \mathbf{H}_1)^N )$.
\end{lemma}

\begin{proof}[{\bf Proof}]
    We first examine the recursion from $t=0$ to $t=N-1$ for each task $m$:
    \begin{align} \notag
        \mathbf{C}_{(m-1)N+t+1} & = (\mathcal{I}-\eta \mathcal{T}_{\mathbf{H}_m} (\eta  ) ) \circ \mathbf{C}_{(m-1)N+t}+\eta^2 \bm{\Sigma}_{\mathbf{H}_m} \\ \notag
        & \preceq   (\mathcal{I}-\eta \widetilde{\mathcal{T}}_{\mathbf{H}_m} (\eta  ) ) \circ \mathbf{C}_{(m-1)N+t} + \eta^2 \mathcal{M}_m \circ \mathbf{C}_{(m-1)N+t}+\eta^2 \sigma^2 \mathbf{H}_m \\ \notag
        & \preceq   (\mathcal{I}-\eta \widetilde{\mathcal{T}}_{\mathbf{H}_m} (\eta  ) ) \circ \mathbf{C}_{(m-1)N+t} + \eta^2 R^2  \cdot \frac{\eta \sigma^2}{1 - \eta R^2} \cdot \mathbf{H}_m +\eta^2 \sigma^2 \mathbf{H}_m \\ \notag
        & \preceq   (\mathcal{I}-\eta \widetilde{\mathcal{T}}_{\mathbf{H}_m} (\eta  ) ) \circ \mathbf{C}_{(m-1)N+t} + \frac{\eta^2 \sigma^2}{1-\eta R^2} \mathbf{H}_m,
    \end{align}
    where the penultimate inequality is derived from the Lemma \ref{lem:c_t_bound}.

    Hence, after $N$ iterations, we could have the following results for task $m$:
    $$
    \mathbf{C}_{(m-1)N+N} \preceq   (\mathcal{I}-\eta \widetilde{\mathcal{T}}_{\mathbf{H}_m} (\eta  ) )^N \circ \mathbf{C}_{(m-1)N} + \frac{\eta^2 \sigma^2}{1-\eta R^2} \sum_{t=0}^{N-1}  (\mathcal{I}-\eta \widetilde{\mathcal{T}}_{\mathbf{H}_m} (\eta  ) )^t \circ \mathbf{H}_m.
    $$
    Now, we consider the first task incorporating with the Lemma B.5 in \cite{zou2021benign}, which implies:
    $$
    \mathbf{C}_N \preceq \frac{\eta \sigma^2}{1-\eta R^2} \cdot (\mathbf{I}-(\mathbf{I}-\eta \mathbf{H}_1)^N ).
    $$
    By combining the aforementioned results and denoting $\mathbf{P}_m =\frac{\eta^2 \sigma^2}{1-\eta R^2} \sum_{t=0}^{N-1}  (\mathcal{I}-\eta \widetilde{\mathcal{T}}_{\mathbf{H}_m}(\eta) )^t \circ \mathbf{H}_m$, we obtain:
    \begin{align}\notag
        \mathbf{C}_{MN} & \preceq \prod_{m=2}^{M}  (\mathcal{I}-\eta \widetilde{\mathcal{T}}_{\mathbf{H}_m} (\eta  ) )^N \circ \mathbf{C}_{N} + \sum_{m=1}^{M-1}\prod_{j=m+1}^M  (\mathcal{I}-\eta \widetilde{\mathcal{T}}_{\mathbf{H}_j} (\eta  ) )^N \mathbf{P}_m + \mathbf{P}_M.
    \end{align}
\end{proof}

Based on Lemma \ref{lem:risk_decom}, the upper bound of the variance error can be expressed as follows:
\begin{align} \notag
     \sum_{k=1}^{M}  \langle \mathbf{H}_k, \mathbf{C}_{MN}  \rangle  &\leq \underbrace{\sum_{k=1}^{M} \frac{\eta \sigma^2}{1-\eta R^2} \langle \mathbf{H}_k, \prod_{m=2}^{M}  (\mathcal{I}-\eta \widetilde{\mathcal{T}}_{\mathbf{H}_m} (\eta  ) )^N  (\mathbf{I}-(\mathbf{I}-\eta \mathbf{H}_1)^N )  \rangle}_{\text{variance term 1}} \\ 
     & + \underbrace{\sum_{k=1}^{M}  \langle \mathbf{H}_k, \sum_{m=1}^{M-1}\prod_{j=m+1}^M  (\mathcal{I}-\eta \widetilde{\mathcal{T}}_{\mathbf{H}_j} (\eta  ) )^N \mathbf{P}_m  \rangle}_{\text{variance term 2}} +  \underbrace{\sum_{k=1}^{M}  \langle \mathbf{H}_k, \mathbf{P}_M  \rangle}_{\text{variance term 3}}.
\end{align}

Let us consider the variance terms separately.
\begin{equation}\label{eq:c_t1_up_1}
    \begin{aligned}
     \text{variance term 1} & = \sum_{k=1}^{M} \frac{\eta \sigma^2}{1-\eta R^2} \langle \mathbf{H}_k, \prod_{m=2}^{M}  (\mathbf{I}-\eta \mathbf{H}_m )^N  (\mathbf{I}-(\mathbf{I}-\eta \mathbf{H}_1)^N )  (\mathbf{I}-\eta \mathbf{H}_m )^N  \rangle \\ 
     & \leq \sum_{k=1}^{M} \frac{\eta \sigma^2}{1-\eta R^2} \langle \prod_{m=2}^{M}  (\mathbf{I}-\eta \mathbf{H}_m )^N \mathbf{H}_k,  (\mathbf{I}-(\mathbf{I}-\eta \mathbf{H}_1)^N )  \rangle \\ 
     & = \frac{\eta \sigma^2}{1-\eta R^2} \sum_{k=1}^{M} \sum_{i}  [ \prod_{m=2}^{M}  (1-\eta \lambda_m^i )^N \lambda_k^i  (1 - (1-\eta \lambda_1^i)^N )  ] \\ 
     & \leq \frac{\eta \sigma^2}{1-\eta R^2} ( \sum_{i<k_1^*} \Gamma_{(2,M)}^i \Lambda^i+ N \eta \sum_{i>k_1^*} \Gamma_{(2,M)}^i  \lambda_1^i\Lambda^i) ,
\end{aligned}
\end{equation}
where we use the facts that $1- (1-\eta \lambda_m^i )^N \leq \min  \{1, \eta N \lambda_m^i \}$ hold for all $i\geq 1$ in the last inequality.

Before we turn our attention to the second term, we first consider the $\mathbf{P}_m$:
\begin{align}\notag
    \mathbf{P}_m &=\frac{\eta^2 \sigma^2}{1-\eta R^2} \sum_{t=0}^{N-1}  (\mathcal{I}-\eta \widetilde{\mathcal{T}}_{\mathbf{H}_m}(\eta) )^t \circ \mathbf{H}_m \\ \notag
    & = \frac{\eta^2 \sigma^2}{1-\eta R^2} \sum_{t=0}^{N-1}  (\mathbf{I}-\eta \mathbf{H}_m )^t \mathbf{H}_m  (\mathbf{I}-\eta \mathbf{H}_m )^t \\\notag
    & \preceq \frac{\eta \sigma^2}{1-\eta R^2}  (\mathbf{I}-(\mathbf{I}-\eta \mathbf{H}_m)^N ).
\end{align}
Substituting the above to the variance term 2, we have:
\begin{align}\notag \label{eq:c_t2_up}
     \text{variance term 2} & =  \sum_{k=1}^{M}  \langle \mathbf{H}_k, \sum_{m=1}^{M-1}\prod_{j=m+1}^M  (\mathcal{I}-\eta \widetilde{\mathcal{T}}_{\mathbf{H}_j} (\eta  ) )^N \mathbf{P}_m  \rangle \\ \notag
     & \leq \frac{\eta \sigma^2}{1-\eta R^2} \sum_{k=1}^{M}  \langle \mathbf{H}_k, \sum_{m=1}^{M-1}\prod_{j=m+1}^M  (\mathcal{I}-\eta \widetilde{\mathcal{T}}_{\mathbf{H}_j} (\eta  ) )^N  (\mathbf{I}-(\mathbf{I}-\eta \mathbf{H}_m)^N )  \rangle \\ \notag 
     & \leq \frac{\eta \sigma^2}{1-\eta R^2} \sum_{k=1}^{M} \langle \mathbf{H}_k, \sum_{m=1}^{M-1}\prod_{j=m+1}^M  (\mathbf{I}-\eta \mathbf{H}_j )^N  (\mathbf{I}-(\mathbf{I}-\eta \mathbf{H}_m)^N )  \rangle \\
     & \leq \sum_{m=1}^{M-1} \frac{\eta \sigma^2}{1-\eta R^2}  ( \sum_{i<k_m^*} \Gamma_{(m+1,M)}^i \Lambda^i+ N \eta \sum_{i>k_m^*} \Gamma_{(m+1,M)}^i  \lambda_m^i\Lambda^i)
\end{align}

Similarly, for the last term, we have:
\begin{align}\label{eq:c_t3_up}
    \text{variance term 3} =  \sum_{k=1}^{M}  \langle \mathbf{H}_k,  \mathbf{P}_M  \rangle 
    \leq \frac{\eta \sigma^2}{1-\eta R^2} ( \sum_{i<k_M^*} \Gamma_{(M+1,M)}^i \Lambda^i+ N \eta \sum_{i>k_M^*} \Gamma_{(M+1,M)}^i  \lambda_m^i\Lambda^i)
\end{align}

\subsection{Lower Bound}
Now, we shift our focus to the lower bound of variance. 
Similarly, we have the following lemma hold:
\begin{lemma}
    Suppose Assumptions \ref{asm:fourth_moment} and \ref{asm:nosie} hold with step size $\eta \leq 1/R^2$, then it holds that:
    \begin{align}\notag
        \mathbf{C}_{MN} & \succeq \prod_{m=2}^{M}  (\mathcal{I}-\eta \widetilde{\mathcal{T}}_{\mathbf{H}_m} (\eta  ) )^N \circ \mathbf{C}_{N} + \sum_{m=1}^{M-1}\prod_{j=m+1}^M  (\mathcal{I}-\eta \widetilde{\mathcal{T}}_{\mathbf{H}_j} (\eta  ) )^N \mathbf{P}_m^{\prime} + \mathbf{P}_M^{\prime},
    \end{align}
    where $\mathbf{P}_m^{\prime} ={\eta^2 \sigma^2} \sum_{t=0}^{N-1}  (\mathcal{I}-\eta \widetilde{\mathcal{T}}_{\mathbf{H}_m}(\eta) )^t \circ \mathbf{H}_m$ and $\mathbf{C}_N \succeq \frac{\eta \sigma^2}{2} \cdot (\mathbf{I}-(\mathbf{I}-\eta \mathbf{H}_1)^{2N} )$.
\end{lemma}
\begin{proof}
In a similar fashion, let's first examine the recursion of $\mathbf{C}$ from $t = 0$ to $t = N-1$ for each task $m$.
\begin{align} \notag
    \mathbf{C}_{(m-1)N+t+1} & = (\mathcal{I}-\eta \mathcal{T}_{\mathbf{H}_m} (\eta  ) ) \circ \mathbf{C}_{(m-1)N+t}+\eta^2 \bm{\Sigma}_{\mathbf{H}_m} \\ \notag
    & =   (\mathcal{I}-\eta \widetilde{\mathcal{T}}_{\mathbf{H}_m} (\eta  ) ) \circ \mathbf{C}_{(m-1)N+t} + \eta^2  (\mathcal{M}_m- \widetilde{\mathcal{M}}_m ) \circ \mathbf{C}_{(m-1)N+t}+\eta^2 \sigma^2 \mathbf{H}_m \\ \notag
    & \succeq   (\mathcal{I}-\eta \widetilde{\mathcal{T}}_{\mathbf{H}_m} (\eta  ) ) \circ \mathbf{C}_{(m-1)N+t}  +\eta^2 \sigma^2 \mathbf{H}_m, 
\end{align}
where we utilize the fact that $\mathcal{M}_m - \widetilde{\mathcal{M}}_m$ is a PSD mapping, as established by \ref{lem:properties}.

Consequently, after $N$ iterations, the following results can be deduced for task $m$:
    $$
    \mathbf{C}_{(m-1)N+N} \succeq   (\mathcal{I}-\eta \widetilde{\mathcal{T}}_{\mathbf{H}_m} (\eta  ) )^N \circ \mathbf{C}_{(m-1)N} + {\eta^2 \sigma^2}\sum_{t=0}^{N-1}  (\mathcal{I}-\eta \widetilde{\mathcal{T}}_{\mathbf{H}_m} (\eta  ) )^t \circ \mathbf{H}_m.
    $$
Now, we consider the first task incorporating the Lemma C.2 in \cite{zou2021benign}, which implies:
    $$
    \mathbf{C}_N \succeq \frac{\eta \sigma^2}{2} \cdot (\mathbf{I}-(\mathbf{I}-\eta \mathbf{H}_1)^{2N} ).
    $$
By combining the aforementioned results and denoting $\mathbf{P}_m^{\prime} = {\eta^2 \sigma^2}  \sum_{t=0}^{N-1}  (\mathcal{I}-\eta \widetilde{\mathcal{T}}_{\mathbf{H}_m}(\eta) )^t \circ \mathbf{H}_m$, we obtain:
\begin{align}\notag
    \mathbf{C}_{MN} & \succeq \prod_{m=2}^{M}  (\mathcal{I}-\eta \widetilde{\mathcal{T}}_{\mathbf{H}_m} (\eta  ) )^N \circ \mathbf{C}_{N} + \sum_{m=1}^{M-1}\prod_{j=m+1}^M  (\mathcal{I}-\eta \widetilde{\mathcal{T}}_{\mathbf{H}_j} (\eta  ) )^N \mathbf{P}_m^{\prime} + \mathbf{P}_M^{\prime},
\end{align}
which completes the proof.
\end{proof}
Drawing from Lemma \ref{lem:risk_decom}, the lower bound of the variance error is expressed as follows:
\begin{align} \notag
     \sum_{k=1}^{M}  \langle \mathbf{H}_k, \mathbf{C}_{MN}  \rangle  &\geq \underbrace{\sum_{k=1}^{M} \frac{\eta \sigma^2}{2} \langle \mathbf{H}_k, \prod_{m=2}^{M}  (\mathcal{I}-\eta \widetilde{\mathcal{T}}_{\mathbf{H}_m} (\eta  ) )^N  (\mathbf{I}-(\mathbf{I}-\eta \mathbf{H}_1)^{2N} )  \rangle}_{\text{variance term 1}^{\prime}} \\ 
     & + \underbrace{\sum_{k=1}^{M}  \langle \mathbf{H}_k, \sum_{m=1}^{M-1}\prod_{j=m+1}^M  (\mathcal{I}-\eta \widetilde{\mathcal{T}}_{\mathbf{H}_j} (\eta  ) )^N \mathbf{P}_m^{\prime}  \rangle}_{\text{variance term 2}^{\prime}} +  \underbrace{\sum_{k=1}^{M}  \langle \mathbf{H}_k, \mathbf{P}_M^{\prime}  \rangle}_{\text{variance term 3}^{\prime}}.
\end{align}

Analogous to the approach for the upper bound, we will examine the terms one by one.
\begin{align}\notag
     \text{variance term 1}^{\prime} & = \sum_{k=1}^{M} \frac{\eta \sigma^2}{2} \langle \mathbf{H}_k, \prod_{m=2}^{M}  (\mathbf{I}-\eta \mathbf{H}_m )^N  (\mathbf{I}-(\mathbf{I}-\eta \mathbf{H}_1)^{2N} )  (\mathbf{I}-\eta \mathbf{H}_m )^N  \rangle \\ \notag
     & = \sum_{k=1}^{M} \frac{\eta \sigma^2}{2} \langle \prod_{m=2}^{M}  (\mathbf{I}-\eta \mathbf{H}_m )^{2N} \mathbf{H}_k,  (\mathbf{I}-(\mathbf{I}-\eta \mathbf{H}_1)^{2N} )  \rangle \\ \notag
     & = \frac{\eta \sigma^2}{2} \sum_{k=1}^{M} \sum_{i}  [ \prod_{m=2}^{M}  (1-\eta \lambda_m^i )^{2N} \lambda_k^i  (1 - (1-\eta \lambda_1^i)^{2N} )  ] \\  \label{eq:c_t1_up_2}
     & \geq \frac{\eta \sigma^2}{2}  \sum_{i}  [ \prod_{m=2}^{M}  (1-\eta \lambda_m^i )^{2N}  ( \sum_{k=1}^{M} \lambda_k^i )  (1 - (1-\eta \lambda_1^i)^{2N} )  ]
\end{align}

To further lower bound the two terms, noticing the following inequality:
$$
1- (1-\eta \lambda_1^i )^{2N} \geq \begin{cases}1- (1-\frac{1}{N} )^{2N} \geq 1-e^{-2} \geq \frac{9}{10}, & \lambda_1^i \geq \frac{1}{\eta N}, \\ 2N \cdot \eta \lambda_1^i-\frac{2N(N-1)}{2} \cdot \eta^2 {\lambda_1^i}^2 \geq \frac{9N}{10} \cdot \eta \lambda_1^i, & \lambda_1^i<\frac{1}{\eta N} .\end{cases}
$$

Hence, the first term, we have:
$$\text{variance term 1}^{\prime} \geq \frac{9 \eta^2 \sigma^2}{20} ( \sum_{i<k_1^*} \Gamma_{(2,M)}^i \Lambda^i+ N \eta \sum_{i>k_1^*} \Gamma_{(2,M)}^i  \lambda_1^i\Lambda^i).$$

For the variance term $2^{\prime}$, we notice that:
\begin{align}\notag
    \mathbf{P}_m^{\prime} &= {\eta^2 \sigma^2}  \sum_{t=0}^{N-1}  (\mathcal{I}-\eta \widetilde{\mathcal{T}}_{\mathbf{H}_m}(\eta) )^t \circ \mathbf{H}_m = {\eta^2 \sigma^2}\sum_{t=0}^{N-1}  (\mathbf{I}-\eta {\mathbf{H}_m}   )^{2t} \mathbf{H}_m \\ \notag
    & \geq  \frac{\eta^2 \sigma^2}{2}  ( \mathbf{I} -  (\mathbf{I}-\eta {\mathbf{H}_m}   )^{2N}  )
\end{align}
Substituting the above to the variance term 2', we have:
\begin{align}\notag
    \text{variance term 2}^{\prime} &= \sum_{k=1}^{M}  \langle \mathbf{H}_k, \sum_{m=1}^{M-1}\prod_{j=m+1}^M  (\mathcal{I}-\eta \widetilde{\mathcal{T}}_{\mathbf{H}_j} (\eta  ) )^N \mathbf{P}_m^{\prime}  \rangle \\ \notag
    & \geq  \frac{\eta^2 \sigma^2}{2}  \sum_{k=1}^{M}  \langle \mathbf{H}_k, \sum_{m=1}^{M-1}\prod_{j=m+1}^M  (\mathcal{I}-\eta \widetilde{\mathcal{T}}_{\mathbf{H}_j} (\eta  ) )^N  ( \mathbf{I} -  (\mathbf{I}-\eta {\mathbf{H}_m}   )^{2N}  )  \rangle \\ \notag
    & = \frac{\eta^2 \sigma^2}{2}  \sum_{k=1}^{M}  \langle \mathbf{H}_k, \sum_{m=1}^{M-1}\prod_{j=m+1}^M  (\mathbf{I}-\eta {\mathbf{H}_j}   )^{2N}   ( \mathbf{I} -  (\mathbf{I}-\eta {\mathbf{H}_m}   )^{2N}  )  \rangle \\ \notag
    & \geq \frac{9 \eta^2 \sigma^2}{20} \sum_{m=1}^{M-1} ( \sum_{i<k_m^*} \Gamma_{(m+1,M)}^i \Lambda^i+ N \eta \sum_{i>k_m^*} \Gamma_{(m+1,M)}^i  \lambda_m^i\Lambda^i).
\end{align}
Also, similar to the variance term 3', it holds that:
$$\text{variance term 3'} \geq \frac{9 \eta^2 \sigma^2}{20} ( \sum_{i<k_M^*} \Gamma_{(M+1,M)}^i \Lambda^i+ N \eta \sum_{i>k_M^*} \Gamma_{(M+1,M)}^i  \lambda_M^i\Lambda^i).$$

\section{Bias Error}
Before providing the proof of bias bound, we first introduce the following lemmas for tradition SGD training in \citealt{zou2021benign}.
\begin{lemma}[Summation of bias iterates \cite{zou2021benign}]
     Suppose that Assumption \ref{asm:fourth_moment} holds. Suppose that $\eta<1 /(\alpha \operatorname{tr}(\mathbf{H}_m))$. Then for every $N \geq 1$ and each task $m$, it holds that:
$$
\frac{1}{2 \eta} \cdot (\mathbf{I}-(\mathbf{I}-\eta \mathbf{H}_m)^{2 N} ) \preceq \sum_{t=0}^{N-1} (\mathcal{I}-\eta \cdot \mathcal{T}_{\mathbf{H}_m}(\eta) )^t \circ \mathbf{H}_m \preceq \frac{1}{\eta} \cdot (\mathbf{I}-(\mathbf{I}-\eta \mathbf{H}_m)^{2 N} )
$$
\end{lemma}


\begin{lemma}\label{lem:s_t bias_accu}
     Under Assumptions \ref{asm:fourth_moment}, let $\mathbf{B}_{a, b}=\mathbf{B}_a-(\mathbf{I}-\eta \mathbf{H}_m)^{b-a} \mathbf{B}_a(\mathbf{I}-\eta \mathbf{H}_m)^{b-a}$, if the stepsize satisfies $\eta<1 /(\alpha_m \operatorname{tr}(\mathbf{H}_m))$, then for any $t \leq N$, it holds that for each task $m$:
$$
\mathbf{S}_t \preceq \sum_{k=0}^{t-1}(\mathbf{I}-\eta \mathbf{H}_m)^k (\frac{\eta \alpha_m \operatorname{tr} (\mathbf{B}_{0, N} )}{1-\eta \alpha_m \operatorname{tr}(\mathbf{H}_m)} \cdot \mathbf{H}_m+\mathbf{B}_0 )(\mathbf{I}-\eta \mathbf{H}_m)^k ,
$$
where denoting $\mathbf{S}_t=\sum_{k=0}^{t-1}(\mathcal{I} - \mathcal{T}(\eta)\circ \mathbf{B}_0$.
\end{lemma}

\begin{lemma}\label{lem:s_t bias_accu_low}
     Suppose Assumptions  \ref{asm:fourth_moment} and \ref{asm:nosie} hold with step size $\eta \leq 1/R^2$, then it holds that:
$$
\mathbf{S}_t \succeq \frac{\beta_m}{4} \operatorname{tr}\left(\left(\mathbf{I}-(\mathbf{I}-\eta \mathbf{H}_m)^{t / 2}\right) \mathbf{B}_0\right) \cdot\left(\mathbf{I}-(\mathbf{I}-\eta \mathbf{H}_m)^{t / 2}\right)+\sum^{t-1}(\mathbf{I}-\eta \mathbf{H}_m)^t \cdot \mathbf{B}_0 \cdot(\mathbf{I}-\eta \mathbf{H}_m)^t,
$$
where denoting $\mathbf{S}_t=\sum_{k=0}^{t-1} (\mathcal{I} - \mathcal{T}(\eta)\circ \mathbf{B}_0$.
\end{lemma}

\subsection{Upper Bound}
\begin{lemma}
    Suppose Assumptions \ref{asm:fourth_moment} and \ref{asm:nosie} hold with step size $\eta \leq 1/R^2$, then it holds that:
    \begin{align}\notag
    \mathbf{B}_{MN} & \preceq \prod_{m=1}^{M}  (\mathcal{I}-\eta \widetilde{\mathcal{T}}_{\mathbf{H}_m} (\eta  ) )^N \circ \mathbf{B}_{0} + \sum_{m=1}^{M}\prod_{j=m}^M  (\mathcal{I}-\eta \widetilde{\mathcal{T}}_{\mathbf{H}_j} (\eta  ) )^N \mathbf{P}_m,
\end{align}
    where $\mathbf{P}_m =\alpha_m\eta^2 \sum_{t=0}^{N-1}  (\mathbf{I}-\eta \mathbf{H}_m )^{2t} \mathbf{H}_m   \langle \mathbf{H}_m, \mathbf{B}_{(m-1)N+t}  \rangle$ and $\prod_{k_1}^{k_2} =1$ if $k_1 > k_2 $.
\end{lemma}

We first examine the recursion from $t=0$ to $t=N-1$ for each task $m$:
\begin{align} \label{eq:b_t_recur_up}
    \mathbf{B}_{(m-1)N+t+1} & = (\mathcal{I}-\eta \mathcal{T}_{\mathbf{H}_m} (\eta  ) ) \circ \mathbf{B}_{(m-1)N+t}  \\ \notag
    & =   (\mathcal{I}-\eta \widetilde{\mathcal{T}}_{\mathbf{H}_m} (\eta  ) ) \circ \mathbf{B}_{(m-1)N+t} + \eta^2  (\mathcal{M}_m - \widetilde{\mathcal{M}}_m )\circ \mathbf{B}_{(m-1)N+t} \\ \notag
    & \preceq   (\mathcal{I}-\eta \widetilde{\mathcal{T}}_{\mathbf{H}_m} (\eta  ) ) \circ \mathbf{B}_{(m-1)N+t} + \alpha_m\eta^2  \cdot \mathbf{H}_m \cdot  \langle \mathbf{H}_m, \mathbf{B}_{(m-1)N+t}  \rangle.  
\end{align}
where the penultimate inequality is derived from the assumption \ref{asm:fourth_moment}.

Hence, after $N$ iterations, we could have the following results for task $m$:
\begin{align} \notag \label{eq:B_m_t}
    \mathbf{B}_{(m-1)N+N}  & \preceq   (\mathcal{I}-\eta \widetilde{\mathcal{T}}_{\mathbf{H}_m} (\eta  ) )^N \circ \mathbf{B}_{(m-1)N} + \alpha_m\eta^2 \sum_{t=0}^{N-1}  (\mathcal{I}-\eta \widetilde{\mathcal{T}}_{\mathbf{H}_m} (\eta  ) )^t \mathbf{H}_m   \langle \mathbf{H}_m, \mathbf{B}_{(m-1)N+t}  \rangle \\ \notag
    & =  (\mathcal{I}-\eta \widetilde{\mathcal{T}}_{\mathbf{H}_m} (\eta  ) )^N \circ \mathbf{B}_{(m-1)N} + \alpha_m\eta^2 \sum_{t=0}^{N-1}  (\mathbf{I}-\eta \mathbf{H}_m )^{2t} \mathbf{H}_m   \langle \mathbf{H}_m, \mathbf{B}_{(m-1)N+t}  \rangle \\ \notag
    & =  (\mathcal{I}-\eta \widetilde{\mathcal{T}}_{\mathbf{H}_m} (\eta  ) )^N \circ \mathbf{B}_{(m-1)N} + \alpha_m\eta^2 \sum_{t=0}^{N-1}  (\mathbf{I}-\eta \mathbf{H}_m )^{2t} \mathbf{H}_m   \langle \mathbf{H}_m,  (\mathcal{I}-\eta {\mathcal{T}}_{\mathbf{H}_m} (\eta  ) )^t \mathbf{B}_{(m-1)N}  \rangle  \\ \notag
    & \preceq  (\mathcal{I}-\eta \widetilde{\mathcal{T}}_{\mathbf{H}_m} (\eta  ) )^N \circ \mathbf{B}_{(m-1)N} + \alpha_m\eta^2 \sum_{t=0}^{N-1} \mathbf{H}_m   \langle \mathbf{H}_m,  (\mathcal{I}-\eta {\mathcal{T}}_{\mathbf{H}_m} (\eta  ) )^t \mathbf{B}_{(m-1)N}  \rangle
\end{align}

We now examine the second term for each $m$:
\begin{equation}\label{eq: bias_up_1}
    \begin{aligned}
        &\sum_{t=0}^{N-1}   \langle \mathbf{H}_m,  (\mathcal{I}-\eta {\mathcal{T}}_{\mathbf{H}_m} (\eta  ) )^t \mathbf{B}_{(m-1)N}  \rangle \\ \notag
      = &\sum_{t=0}^{N-1}   \langle \mathbf{H}_m,  (\mathcal{I}-\eta {\mathcal{T}}_{\mathbf{H}_m} (\eta  ) )^t  (\mathcal{I}-\eta {\mathcal{T}}_{\mathbf{H}_{m-1}} (\eta  ) )^N \ldots  (\mathcal{I}-\eta {\mathcal{T}}_{\mathbf{H}_{1}} (\eta  ) )^N \mathbf{B}_{0}  \rangle\\ \notag
      =& \sum_{t=0}^{N-1}   \langle  (\mathcal{I}-\eta {\mathcal{T}}_{\mathbf{H}_{m-1}} (\eta  ) )^N \ldots  (\mathcal{I}-\eta {\mathcal{T}}_{\mathbf{H}_{1}} (\eta  ) )^N \mathbf{H}_m,  (\mathcal{I}-\eta {\mathcal{T}}_{\mathbf{H}_m} (\eta  ) )^t \mathbf{B}_{0}  \rangle,
    \end{aligned}
\end{equation}
where we know the following holds:
\begin{align*}
      (\mathcal{I}-\eta {\mathcal{T}}_{\mathbf{H}_{m-1}} (\eta  ) ) \circ \mathbf{H}_m & =  (\mathcal{I}-\eta \widetilde{\mathcal{T}}_{\mathbf{H}_{m-1}} (\eta  ) ) \circ \mathbf{H}_m +  (\mathcal{M} -\widetilde{\mathcal{M}} )\circ \mathbf{H}_m\\
     & \preceq    (\mathcal{I}-\eta \widetilde{\mathcal{T}}_{\mathbf{H}_{m-1}} (\eta  ) ) \circ \mathbf{H}_m  + \alpha_{m-1}\eta^2  \cdot \mathbf{H}_{m-1} \cdot  \langle \mathbf{H}_{m-1}, \mathbf{H}_{m}  \rangle.  
\end{align*}
Moreover, we have $\sum_{t=0}^{N-1} \eta \cdot \mathbf{H}_{m-1} \cdot (\mathcal{I}-\eta \widetilde{\mathcal{T}}_{\mathbf{H}_{m-1}}(\eta ))^t \preceq \mathbf{I} - (\mathbf{I}- \eta \mathbf{H}_{m-1})^N \preceq \mathbf{I} $. Therefore, it holds that:
\begin{align*}
      (\mathcal{I}-\eta {\mathcal{T}}_{\mathbf{H}_{m-1}} (\eta  ) )^N \circ \mathbf{H}_m \preceq    (\mathcal{I}-\eta \widetilde{\mathcal{T}}_{\mathbf{H}_{m-1}} (\eta  ) )^N \circ \mathbf{H}_m  + \alpha_{m-1}\eta \cdot \mathbf{I} \cdot  \langle \mathbf{H}_{m-1}, \mathbf{H}_{m}  \rangle.  
\end{align*}
It implies:
\begin{align*}
      (\mathcal{I}-\eta {\mathcal{T}}_{\mathbf{H}_{1}} (\eta  ) )^N \ldots  (\mathcal{I}-\eta {\mathcal{T}}_{\mathbf{H}_{m-1}} (\eta  ) )^N \circ \mathbf{H}_m &\preceq   (\mathcal{I}-\eta \widetilde{\mathcal{T}}_{\mathbf{H}_{1}} (\eta  ) )^N \ldots  (\mathcal{I}-\eta \widetilde{\mathcal{T}}_{\mathbf{H}_{m-1}} (\eta  ) )^N \circ \mathbf{H}_m\\
     &+ \sum_{j=1}^{m-1} \prod_{k=1}^{j} \alpha_{k}\eta^j  \cdot \langle \mathbf{H}_{k-1}, \mathbf{I} - (\mathbf{I}- \eta \mathbf{H}_{m-1})^N \rangle \cdot  \langle \mathbf{H}_{j}, \mathbf{H}_{m}  \rangle \cdot \mathbf{I},
\end{align*}
where we denote $\mathbf{H}_0 = \mathbf{I} $ and define $\Phi_1^{m-1} := \sum_{j=1}^{m-1} \prod_{k=1}^{j} \alpha_{k}\eta^j  \cdot \langle \mathbf{H}_{k-1}, \mathbf{I} - (\mathbf{I}- \eta \mathbf{H}_{m-1})^N \rangle \cdot  \langle \mathbf{H}_{j}, \mathbf{H}_{m}  \rangle$. Therefore, \cref{eq: bias_up_1} can be represented as follows:
{\small
\begin{align*}\notag 
     &\sum_{t=0}^{N-1}   \langle  (\mathcal{I}-\eta \widetilde{\mathcal{T}}_{\mathbf{H}_{1}} (\eta  ) )^N \ldots  (\mathcal{I}-\eta \widetilde{\mathcal{T}}_{\mathbf{H}_{m-1}} (\eta  ) )^N \circ \mathbf{H}_m +{\Phi_1^{m-1}} \cdot \mathbf{I},  (\mathcal{I}-\eta {\mathcal{T}}_{\mathbf{H}_m} (\eta  ) )^t \mathbf{B}_{0}  \rangle  \\
     \leq &  \underbrace{\sum_{t=0}^{N-1}   \langle  (\mathcal{I}-\eta \widetilde{\mathcal{T}}_{\mathbf{H}_{1}} (\eta  ) )^N \ldots  (\mathcal{I}-\eta \widetilde{\mathcal{T}}_{\mathbf{H}_{m-1}} (\eta  ) )^N \circ \mathbf{H}_m , (\mathbf{I}-\eta \mathbf{H}_m)^t  (\frac{\eta \alpha_m \operatorname{tr} (\mathbf{B}_{0, N} )}{1-\eta \alpha_m \operatorname{tr}(\mathbf{H}_m)} \cdot \mathbf{H}_m+\mathbf{B}_0 )(\mathbf{I}-\eta \mathbf{H}_m)^t  \rangle}_{\text{term 1}} \\
     +& \underbrace{\sum_{t=0}^{N-1}   \langle{\Phi_1^{m-1}} \cdot \mathbf{I} , (\mathbf{I}-\eta \mathbf{H}_m)^t  (\frac{\eta \alpha_m \operatorname{tr} (\mathbf{B}_{0, N} )}{1-\eta \alpha_m \operatorname{tr}(\mathbf{H}_m)} \cdot \mathbf{H}_m+\mathbf{B}_0 )(\mathbf{I}-\eta \mathbf{H}_m)^t  \rangle }_{\text{term 2}}.
\end{align*}}
We first consider the term 1 with \cref{lem:s_t bias_accu}.
{\small
\begin{align*}
    \text{term 1} =  &  \sum_{t=0}^{N-1}   \langle \prod_{j=1}^{m-1}(\mathbf{I}- \eta \mathbf{H}_{j})^{2N} (\mathbf{I}-\eta \mathbf{H}_m)^{2t} \mathbf{H}_m,  (\frac{\eta \alpha_m \operatorname{tr} (\mathbf{B}_{0, N} )}{1-\eta \alpha_m \operatorname{tr}(\mathbf{H}_m)} \cdot \mathbf{H}_m+\mathbf{B}_0 )  \rangle \\ \notag 
      = & \sum_{t=0}^{N-1} \frac{\eta \alpha_m \operatorname{tr} (\mathbf{B}_{0, N} )}{1-\eta \alpha_m \operatorname{tr}(\mathbf{H}_m)}  \langle \prod_{j=1}^{m-1}(\mathbf{I}- \eta \mathbf{H}_{j})^{2N}(\mathbf{I}-\eta \mathbf{H}_m)^{2t} \mathbf{H}_m,  \mathbf{H}_m \rangle + \sum_{t=0}^{N-1}  \langle (\mathbf{I}-\eta \mathbf{H}_m)^{2t} \mathbf{H}_m, \mathbf{B}_0 \rangle \\ \notag 
      \leq & \frac{ \alpha_m \operatorname{tr} (\mathbf{B}_{0, N} )}{1-\eta \alpha_m \operatorname{tr}(\mathbf{H}_m)}  \langle \prod_{j=1}^{m-1}(\mathbf{I}- \eta \mathbf{H}_{j})^{2N} (\mathbf{I} - ( \mathbf{I} - \eta\mathbf{H}_m)^N) , \mathbf{H}_m  \rangle + \frac{1}{\eta} \langle \prod_{j=1}^{m-1}(\mathbf{I}- \eta \mathbf{H}_{j})^{2N}(\mathbf{I} - ( \mathbf{I} - \eta\mathbf{H}_m)^N) , \mathbf{B}_0  \rangle \\ \notag
       =& \frac{ \alpha_m \operatorname{tr} (\mathbf{B}_{0, N} )}{1-\eta \alpha_m \operatorname{tr}(\mathbf{H}_m)} \sum_i \Gamma_{(1,m-1)}^i [1- (1-\eta \lambda_m^i  )^N ]\lambda_m^i + \frac{1}{\eta}\sum_i \Gamma_{(1,m-1)}^i{\omega_i^2} [1- (1-\eta \lambda_m^i )^N ] \\ \notag
       \leq & \frac{ \alpha_m \operatorname{tr} (\mathbf{B}_{0, N} )}{1-\eta \alpha_m \operatorname{tr}(\mathbf{H}_m)} \sum_i \Gamma_{(1,m-1)}^i\min  \{1, \eta N \lambda_m^i \} + \frac{1}{\eta}\sum_i\Gamma_{(1,m-1)}^i {\omega_i^2} \min  \{1, \eta N \lambda_m^i \} \\ \notag
       \leq  & \frac{ \alpha_m \operatorname{tr} (\mathbf{B}_{0, N} )}{1-\eta \alpha_m \operatorname{tr}(\mathbf{H}_m)} (\sum_{i\leq k_m^* }\frac{\Gamma_{(1,m-1)}^i \lambda_m^i}{N \eta}+N \eta \sum_{i>k_m^*} \Gamma_{(1,m-1)}^i(\lambda_m^i)^2 ) + \frac{1}{\eta} \|\mathbf{w}_0-\mathbf{w}^* \|_{\bm{\Gamma}_1^{m-1}\mathbf{I}_{m, 0: k_m^*}}^2+ N  \|\mathbf{w}_0-\mathbf{w}^* \|_{\bm{\Gamma}_1^{m-1}\mathbf{H}_{m, k_m^*: \infty}}^2
\end{align*}}
\hspace{-0.12cm}where $k_m^*$ is the index of the smallest eigenvalue of $\mathbf{H}_m$ satisfying $\lambda_{k_m^*}^i \geq 1 / (\eta N)$, and denotes $U_m = \frac{ \alpha_m \operatorname{tr} (\mathbf{B}_{0, N} )}{1-\eta \alpha_m \operatorname{tr}(\mathbf{H}_m)} (\sum_{i\leq k_m^* }\frac{\Gamma_{(1,m-1)}^i}{N \eta}+N \eta \sum_{i>k_m^*} (\lambda_m^i)^2 ) + \frac{1}{\eta} \|\mathbf{w}_0-\mathbf{w}^* \|_{\bm{\Gamma}_1^{m-1}\mathbf{I}_{m, 0: k_m^*}}^2+ N  \|\mathbf{w}_0-\mathbf{w}^* \|_{\bm{\Gamma}_1^{m-1}\mathbf{H}_{m, k_m^*: \infty}}^2$.

Moreover, $ .\operatorname{tr} (\mathbf{B}_{0, N} )=\operatorname{tr} (\mathbf{B}_0-(\mathbf{I}-\eta \mathbf{H}_m)^N \mathbf{B}_0(\mathbf{I}-\eta \mathbf{H}_m)^N ) )=\sum (1- (1-\eta \Lambda^i)^{2 N} ) \cdot ( \langle\mathbf{w}_0-\mathbf{w}^*, \mathbf{v}_i \rangle )^2$, Hence:
$$
\operatorname{tr} (\mathbf{B}_{0, N} ) \leq 2 \sum_i \min  \{1, N \eta \Lambda^i\} ( \langle\mathbf{w}_0-\mathbf{w}^*, \mathbf{v}_i \rangle )^2 \leq 2 ( \|\mathbf{w}_0-\mathbf{w}^* \|_{\mathbf{I}_{m, 0: {k_m^*}}}^2+N \eta \|\mathbf{w}_0-\mathbf{w}^* \|_{\mathbf{H}_{m, {k_m^*}: \infty}}^2 ).
$$

Now we are ready to examine the term 2.
\begin{align*}
    &\text{term 2} =    \sum_{t=0}^{N-1}   \langle{\Phi_1^{m-1}} \cdot \mathbf{I}  \cdot  (\mathbf{I}-\eta \mathbf{H}_m)^{2t},  (\frac{\eta \alpha_m \operatorname{tr} (\mathbf{B}_{0, N} )}{1-\eta \alpha_m \operatorname{tr}(\mathbf{H}_m)} \cdot \mathbf{H}_m+\mathbf{B}_0 )  \rangle \\ \notag 
    &\leq  \sum_{t=0}^{N-1} \frac{\eta \alpha_m \operatorname{tr} (\mathbf{B}_{0, N} )}{1-\eta \alpha_m \operatorname{tr}(\mathbf{H}_m)}  \langle {\Phi_1^{m-1}} \cdot (\mathbf{I}-\eta \mathbf{H}_m)^{2t} ,  \mathbf{H}_m \rangle + \sum_{t=0}^{N-1}  \langle  {\Phi_1^{m-1}} \cdot (\mathbf{I}-\eta \mathbf{H}_m)^{2t}, \mathbf{B}_0 \rangle \\ \notag 
    &\leq  \frac{ \alpha_m \operatorname{tr} (\mathbf{B}_{0, N} )}{1-\eta \alpha_m \operatorname{tr}(\mathbf{H}_m)}  \langle {\Phi_1^{m-1}} \mathbf{I}  , (\mathbf{I} - ( \mathbf{I} - \eta\mathbf{H}_m)^N)  \rangle  + \frac{1}{\eta} \langle \sum_{j=1}^{m-1}  \langle \mathbf{H}_{j}, \alpha^j \bm{\Gamma}_1^{j-1} \cdot \mathbf{H}_m \rangle \mathbf{H}_m^{-1} (\mathbf{I} - ( \mathbf{I} - \eta\mathbf{H}_m)^N) , \mathbf{B}_0  \rangle \\ \notag
    &= \frac{ \alpha_m \operatorname{tr} (\mathbf{B}_{0, N} )}{1-\eta \alpha_m \operatorname{tr}(\mathbf{H}_m)} \sum_i {\Phi_1^{m-1}} [1- (1-\eta \lambda_m^i  )^N ] + \frac{1}{\eta}\sum_i {\Phi_1^{m-1}}^i{\omega_i^2} (\lambda_m^i)^{-1} [1- (1-\eta \lambda_m^i )^N ] \\ \notag
    &\leq  \frac{ \alpha_m \operatorname{tr} (\mathbf{B}_{0, N} )}{1-\eta \alpha_m \operatorname{tr}(\mathbf{H}_m)} \sum_i {\Phi_1^{m-1}}\min  \{1, \eta N \lambda_m^i \} + \frac{1}{\eta}\sum_i{\Phi_1^{m-1}} {\omega_i^2} (\lambda_m^i)^{-1} \min  \{1, \eta N \lambda_m^i \} \\ \notag
     &\leq  \frac{ \alpha_m \operatorname{tr} (\mathbf{B}_{0, N} ) {\Phi_1^{m-1}}}{1-\eta \alpha_m \operatorname{tr}(\mathbf{H}_m)} (k_m^*+N \eta \sum_{i>k_m^*} (\lambda_m^i) ) + \frac{{\Phi_1^{m-1}}}{\eta} \|\mathbf{w}_0-\mathbf{w}^* \|_{\mathbf{H}^{-1}_{m, 0: k_m^*}}^2+ N{\Phi_1^{m-1}} \|\mathbf{w}_0-\mathbf{w}^* \|_{\mathbf{I}_{m, k_m^*: \infty}}^2.
\end{align*}
Let us denote $V_m = \frac{ \alpha_m \operatorname{tr} (\mathbf{B}_{0, N} ) {\Phi_1^{m-1}}}{1-\eta \alpha_m \operatorname{tr}(\mathbf{H}_m)} (k_m^*+N \eta \sum_{i>k_m^*} (\lambda_m^i) ) + \frac{{\Phi_1^{m-1}}}{\eta} \|\mathbf{w}_0-\mathbf{w}^* \|_{\mathbf{H}^{-1}_{m, 0: k_m^*}}^2+ N{\Phi_1^{m-1}} \|\mathbf{w}_0-\mathbf{w}^* \|_{\mathbf{I}_{m, k_m^*: \infty}}^2$.

By combining the aforementioned results, we obtain:
\begin{align}\notag
    \mathbf{B}_{MN} & \preceq \prod_{m=1}^{M}  (\mathcal{I}-\eta \widetilde{\mathcal{T}}_{\mathbf{H}_m} (\eta  ) )^N \circ \mathbf{B}_{0} + \sum_{m=1}^{M}\prod_{j=m}^M  (\mathcal{I}-\eta \widetilde{\mathcal{T}}_{\mathbf{H}_j} (\eta  ) )^N \mathbf{P}_m,
\end{align}
where denoting $\mathbf{P}_m =\alpha_m\eta^2 (U_m +V_m)\cdot \mathbf{H}_m $.

Based on Lemma \ref{lem:risk_decom}, the upper bound of the bias error can be expressed as follows:
\begin{align} \notag
     \sum_{k=1}^{M}  \langle \mathbf{H}_k, \mathbf{B}_{MN}  \rangle  \leq \underbrace{\sum_{k=1}^{M}   \langle \mathbf{H}_k, \prod_{m=1}^{M}  (\mathcal{I}-\eta \widetilde{\mathcal{T}}_{\mathbf{H}_m} (\eta  ) )^N \circ \mathbf{B}_{0}  \rangle}_{\text{bias term 1}}  + \underbrace{\sum_{k=1}^{M}  \langle \mathbf{H}_k, \sum_{m=1}^{M}\prod_{j=m}^M  (\mathcal{I}-\eta \widetilde{\mathcal{T}}_{\mathbf{H}_j} (\eta  ) )^N \mathbf{P}_m  \rangle}_{\text{bias term 2}}. 
\end{align}

For each $k$:
\begin{align}\notag
     & \langle \mathbf{H}_k, \mathbf{B}_{MN}  \rangle  =    \langle \prod_{m=1}^{M}  (\mathbf{I}-\eta {\mathbf{H}_m} )^{2N}\mathbf{H}_k, \mathbf{B}_{0}  \rangle +  \langle \mathbf{H}_k, \sum_{m=1}^{M}\prod_{j=m}^M  (\mathbf{I}-\eta {\mathbf{H}_j} )^{2N} \alpha_m\eta^2 (U_m +V_m) \cdot \mathbf{H}_m  \rangle \\ \notag
     & \leq  \|\mathbf{w}_0-\mathbf{w}^* \|_{\prod_{m=1}^{M}  (\mathbf{I}-\eta {\mathbf{H}_m} )^{2N}\mathbf{H}_k}^2 \\ \notag
     & +  \sum_{m=1}^M \alpha_m \eta^2  ( \frac{ \alpha_m  \operatorname{tr} (\mathbf{B}_{0, N} )}{1-\eta \alpha_m \operatorname{tr}(\mathbf{H}_m)} (\sum_{i<k_m^*} \frac{\Gamma_{(1,M)}^i(\lambda_m^i)^2 \lambda_k^i}{N \eta}+N \eta \sum_{i>k_m^*} \Gamma_{(1,M)}^i (\lambda_m^i)^3 \lambda_k^i )  ) \\
     &+   \sum_{m=1}^M \alpha_m \eta^2 (\|\mathbf{w}_0-\mathbf{w}^* \|_{(\bm{\Gamma}_1^M\mathbf{H}_m\mathbf{H}_k)_{ 0: k_m^*}}^2+ N \eta \|\mathbf{w}_0-\mathbf{w}^* \|_{(\bm{\Gamma}_1^M\mathbf{H}_m^2\mathbf{H}_k)_{ k_m^*: \infty}}^2  ) \\ \notag
     & + \sum_{m=1}^M \alpha_m \eta^2 {\Phi_1^{m-1}}(\frac{ \alpha_m \operatorname{tr} (\mathbf{B}_{0, N} ) }{1-\eta \alpha_m \operatorname{tr}(\mathbf{H}_m)} (\sum_{i<k_m^*} \Gamma_{(m,M)}^{i} (\lambda_m^i)+N \eta \sum_{i>k_m^*} \Gamma_{(m,M)}^{i} (\lambda_m^i)^2 )) \\ \notag
     &+ \sum_{m=1}^M \alpha_m \eta^2 {\Phi_1^{m-1}}(\frac{1}{\eta} \|\mathbf{w}_0-\mathbf{w}^* \|_{(\bm{\Gamma}_m^M\mathbf{H}^{-1}_{m}){ 0: k_m^*}}^2+ N\|\mathbf{w}_0-\mathbf{w}^* \|_{(\bm{\Gamma}_m^M )_{ k_m^*: \infty}}^2 )
\end{align}
Hence,
\begin{align*} \notag
     \sum_{k=1}^{M}  \langle \mathbf{H}_k, \mathbf{B}_{MN}  \rangle & \leq \sum_{k=1}^{M}   \|\mathbf{w}_0-\mathbf{w}^* \|_{\prod_{m=1}^{M}  (\mathbf{I}-\eta {\mathbf{H}_m} )^{2N}\mathbf{H}_k}^2  \\
     &+ \sum_{k=1}^{M}  \sum_{m=1}^M \alpha_m \eta^2 \frac{ \alpha_m \operatorname{tr} (\mathbf{B}_{0, N} )}{1-\eta \alpha_m \operatorname{tr}(\mathbf{H}_m)} (\sum_{i<k_m^*} \frac{\Gamma_{(1,M)}^i(\lambda_m^i)^2\lambda_k^i}{N \eta}+N \eta \sum_{i>k_m^*} \Gamma_{(1,M)}^i (\lambda_m^i)^3\lambda_k^i ) \\
     & + \sum_{k=1}^{M}    \sum_{m=1}^M \alpha_m \eta^2  (  \|\mathbf{w}_0-\mathbf{w}^* \|_{(\bm{\Gamma}_1^{M}\mathbf{H}_m\mathbf{H}_k)_{ 0: k_m^*}}^2+ N \eta \|\mathbf{w}_0-\mathbf{w}^* \|_{{(\bm{\Gamma}_1^{M}\mathbf{H}_k\mathbf{H}_m^2)}_{ k_m^*: \infty}}^2  ) \\
     & + \sum_{k=1}^{M}\sum_{m=1}^M \alpha_m \eta^2 {\Phi_1^{m-1}}(\frac{ \alpha_m \operatorname{tr} (\mathbf{B}_{0, N} ) }{1-\eta \alpha_m \operatorname{tr}(\mathbf{H}_m)} (\sum_{i<k_m^*} \Gamma_{(m,M)}^{i}\lambda_k^i (\lambda_m^i)+N \eta \sum_{i>k_m^*} \Gamma_{(m,M)}^{i} (\lambda_m^i)^2 \lambda_k^i )) \\ \notag
     &+ \sum_{k=1}^{M}\sum_{m=1}^M \alpha_m \eta {\Phi_1^{m-1}}( \|\mathbf{w}_0-\mathbf{w}^* \|_{(\bm{\Gamma}_m^M \mathbf{H}_k){ 0: k_m^*}}^2+ N\|\mathbf{w}_0-\mathbf{w}^* \|_{(\bm{\Gamma}_m^M \mathbf{H}_k\mathbf{H}_{m})_{ k_m^*: \infty}}^2 )
\end{align*}

\subsection{Lower Bound}
We first examine the recursion from $t=0$ to $t=N-1$ for each task $m$:
\begin{align} 
    \mathbf{B}_{(m-1)N+t+1} & = (\mathcal{I}-\eta \mathcal{T}_{\mathbf{H}_m} (\eta  ) ) \circ \mathbf{B}_{(m-1)N+t}  \\ \notag
    & =   (\mathcal{I}-\eta \widetilde{\mathcal{T}}_{\mathbf{H}_m} (\eta  ) ) \circ \mathbf{B}_{(m-1)N+t} + \eta^2  (\mathcal{M}_m - \widetilde{\mathcal{M}}_m )\circ \mathbf{B}_{(m-1)N+t} \\ \notag
    & \succeq   (\mathcal{I}-\eta \widetilde{\mathcal{T}}_{\mathbf{H}_m} (\eta  ) ) \circ \mathbf{B}_{(m-1)N+t} + \beta_m \eta^2 \cdot \mathbf{H}_m \cdot  \langle \mathbf{H}_m, \mathbf{B}_{(m-1)N+t}  \rangle. 
\end{align}
Hence, after $N$ iterations, we could have the following results for task $m$:
\begin{align} \notag 
    \mathbf{B}_{(m-1)N+N}  & \succeq   (\mathcal{I}-\eta \widetilde{\mathcal{T}}_{\mathbf{H}_m} (\eta  ) )^N \circ \mathbf{B}_{(m-1)N} + \beta_m\eta^2 \sum_{t=0}^{N-1}  (\mathcal{I}-\eta \widetilde{\mathcal{T}}_{\mathbf{H}_m} (\eta  ) )^t \mathbf{H}_m   \langle \mathbf{H}_m, \mathbf{B}_{(m-1)N+t}  \rangle \\ \notag
    & =  (\mathcal{I}-\eta \widetilde{\mathcal{T}}_{\mathbf{H}_m} (\eta  ) )^N \circ \mathbf{B}_{(m-1)N} + \beta_m\eta^2 \sum_{t=0}^{N-1}  (\mathbf{I}-\eta \mathbf{H}_m )^{2t} \mathbf{H}_m   \langle \mathbf{H}_m, \mathbf{B}_{(m-1)N+t}  \rangle \\ \notag
    & =  (\mathcal{I}-\eta \widetilde{\mathcal{T}}_{\mathbf{H}_m} (\eta  ) )^N \circ \mathbf{B}_{(m-1)N} + \beta_m\eta^2 \sum_{t=0}^{N-1}  (\mathbf{I}-\eta \mathbf{H}_m )^{2t} \mathbf{H}_m   \langle \mathbf{H}_m,  (\mathcal{I}-\eta {\mathcal{T}}_{\mathbf{H}_m} (\eta  ) )^t \mathbf{B}_{(m-1)N}  \rangle \\ 
\end{align}

We now examine the second term for each $m$:
\begin{equation}\label{eq: low_bias_15}
    \begin{aligned}
        &\beta_m\eta^2 \sum_{t=0}^{N-1}  (\mathbf{I}-\eta \mathbf{H}_m )^{2t} \mathbf{H}_m   \langle \mathbf{H}_m,  (\mathcal{I}-\eta {\mathcal{T}}_{\mathbf{H}_m} (\eta  ) )^t \mathbf{B}_{(m-1)N}  \rangle \\ 
      = \quad &\beta_m\eta^2 \sum_{t=0}^{N-1}  (\mathbf{I}-\eta \mathbf{H}_m )^{2t} \mathbf{H}_m    \langle \mathbf{H}_m,  (\mathcal{I}-\eta {\mathcal{T}}_{\mathbf{H}_m} (\eta  ) )^t  (\mathcal{I}-\eta {\mathcal{T}}_{\mathbf{H}_{m-1}} (\eta  ) )^N \ldots  (\mathcal{I}-\eta {\mathcal{T}}_{\mathbf{H}_{1}} (\eta  ) )^N \mathbf{B}_{0}  \rangle 
    \end{aligned}
\end{equation}

Acccording to \cref{asm:fourth_moment} \ref{asm:fourth_lower}, we have:
\begin{align*}
      (\mathcal{I}-\eta {\mathcal{T}}_{\mathbf{H}_{m-1}} (\eta  ) ) \circ \mathbf{H}_m & =  (\mathcal{I}-\eta \widetilde{\mathcal{T}}_{\mathbf{H}_{m-1}} (\eta  ) ) \circ \mathbf{H}_m +  (\mathcal{M} -\widetilde{\mathcal{M}} )\circ \mathbf{H}_m\\
     & \succeq    (\mathcal{I}-\eta \widetilde{\mathcal{T}}_{\mathbf{H}_{m-1}} (\eta  ) ) \circ \mathbf{H}_m  + \beta_{m-1}\eta^2  \cdot \mathbf{H}_{m-1} \cdot  \langle \mathbf{H}_{m-1}, \mathbf{H}_{m}  \rangle\\
     \rightarrow (\mathcal{I}-\eta {\mathcal{T}}_{\mathbf{H}_{m-1}} (\eta  ) )^N \circ \mathbf{H}_m & \succeq (\mathcal{I}-\eta \widetilde{\mathcal{T}}_{\mathbf{H}_{m-1}} (\eta  ) )^N \circ \mathbf{H}_m + \beta_{m-1}\eta^2 \cdot  \sum_{t=0}^{N-1} (\mathcal{I}-\eta \widetilde{\mathcal{T}}_{\mathbf{H}_{m-1}} (\eta  ) )^t \circ \mathbf{H}_{m-1} \cdot  \langle \mathbf{H}_{m-1}, \mathbf{H}_{m}  \rangle \\
     & \succeq (\mathcal{I}-\eta \widetilde{\mathcal{T}}_{\mathbf{H}_{m-1}} (\eta  ) )^N \circ \mathbf{H}_m + \frac{\beta_{m-1}\eta}{2} \cdot  ( \mathbf{I}-(\mathbf{I}-\eta {\mathbf{H}_{m-1}})^{2N}) \cdot  \langle \mathbf{H}_{m-1}, \mathbf{H}_{m}  \rangle.
\end{align*}
Therefore, we have iterations that:
\begin{align*}
      (\mathcal{I}-\eta {\mathcal{T}}_{\mathbf{H}_{1}} (\eta  ) )^N \ldots  (\mathcal{I}-\eta {\mathcal{T}}_{\mathbf{H}_{m-1}} (\eta  ) )^N \circ \mathbf{H}_m &\succeq   (\mathcal{I}-\eta \widetilde{\mathcal{T}}_{\mathbf{H}_{1}} (\eta  ) )^N \ldots  (\mathcal{I}-\eta \widetilde{\mathcal{T}}_{\mathbf{H}_{m-1}} (\eta  ) )^N \circ \mathbf{H}_m\\
     &+ \sum_{j=1}^{m-1} \prod_{k=1}^{j} \beta_{k} (\frac{\eta}{2})^j  \cdot \langle\mathbf{H}_{k-1},( \mathbf{I}-(\mathbf{I}-\eta {\mathbf{H}_{m-1}})^{2N}) \rangle \cdot  \langle \mathbf{H}_{j}, \mathbf{H}_{m}  \rangle \cdot \mathbf{I}.
\end{align*}

Subsituting the above to \cref{eq: low_bias_15} and denoting $\hat{\Phi}_1^{m-1}:=\sum_{j=1}^{m-1} \prod_{k=1}^{j} \beta_{k} (\frac{\eta}{2})^j  \cdot \langle\mathbf{H}_{k-1},( \mathbf{I}-(\mathbf{I}-\eta {\mathbf{H}_{m-1}})^{2N}) \rangle \cdot  \langle \mathbf{H}_{j}, \mathbf{H}_{m}  \rangle$, we have:
\begin{align*}
   \sum_{t=0}^{N-1} \langle \mathbf{H}_m,  (\mathcal{I}-\eta {\mathcal{T}}_{\mathbf{H}_m} (\eta  ) )^t \mathbf{B}_{(m-1)N}  \rangle  &\succeq     \sum_{t=0}^{N-1}\langle  (\mathcal{I}-\eta {\widetilde{\mathcal{T}}}_{\mathbf{H}_{m-1}} (\eta  ) )^N \ldots  (\mathcal{I}-\eta {\widetilde{\mathcal{T}}}_{\mathbf{H}_{1}} (\eta  ) )^N  \mathbf{H}_m,  (\mathcal{I}-\eta {\mathcal{T}}_{\mathbf{H}_m} (\eta  ) )^t \mathbf{B}_{0}  \rangle \\
    &+ \sum_{t=0}^{N-1}\langle \hat{\Phi}_1^{m-1}\mathbf{I} , (\mathcal{I}-\eta {\mathcal{T}}_{\mathbf{H}_m} (\eta  ) )^t \mathbf{B}_{0} \rangle \\
    & =   \underbrace{\langle \prod_{p=1}^{m-1} (\mathbf{I}-\eta \mathbf{H}_p )^{2N}  \mathbf{H}_m, \sum_{t=0}^{N-1}  (\mathcal{I}-\eta {\mathcal{T}}_{\mathbf{H}_m} (\eta  ) )^t \mathbf{B}_{0}  \rangle}_{\text{term 1}}  \\
    &+  \underbrace{\langle \hat{\Phi}_1^{m-1}\mathbf{I} ,\sum_{t=0}^{N-1} (\mathcal{I}-\eta {\mathcal{T}}_{\mathbf{H}_m} (\eta  ) )^t \mathbf{B}_{0} \rangle }_{\text{term 2}} 
\end{align*}

From the Lemma, we have:
\begin{align} \notag
    \sum_{t=0}^{N-1}  (\mathcal{I}-\eta {\mathcal{T}}_{\mathbf{H}_m} (\eta  ) )^t \mathbf{B}_{0} &\succeq \frac{\beta_m}{4} \operatorname{tr} ( (\mathbf{I}-(\mathbf{I}-\eta \mathbf{H}_m)^{N / 2} ) \mathbf{B}_0 ) \cdot (\mathbf{I}-(\mathbf{I}-\eta \mathbf{H}_m)^{N / 2} ) \\ \notag 
    &+\sum_{t=0}^{N-1}(\mathbf{I}-\eta \mathbf{H}_m)^t \cdot \mathbf{B}_0 \cdot(\mathbf{I}-\eta \mathbf{H}_m)^t .    
\end{align}

Then, for each task $m$, we examine the term 1:
\begin{align}\notag
    \text{term 1} =& \langle \prod_{p=1}^{m-1} (\mathbf{I}-\eta \mathbf{H}_p )^{2N}  \mathbf{H}_m, \sum_{t=0}^{N-1}  (\mathcal{I}-\eta {\mathcal{T}}_{\mathbf{H}_m} (\eta  ) )^t \mathbf{B}_{0}  \rangle \\ \notag 
    \geq &  \langle \prod_{p=1}^{m-1} (\mathbf{I}-\eta \mathbf{H}_p )^{2N}  \mathbf{H}_m,  \frac{\beta_m}{4} \operatorname{tr} ( (\mathbf{I}-(\mathbf{I}-\eta \mathbf{H}_m)^{N / 2} ) \mathbf{B}_0 ) \cdot (\mathbf{I}-(\mathbf{I}-\eta \mathbf{H}_m)^{N / 2} )  \rangle \\ \notag
    + &   \langle \prod_{p=1}^{m-1} (\mathbf{I}-\eta \mathbf{H}_p )^{2N}  \mathbf{H}_m,  \sum_{t=0}^{N-1}(\mathbf{I}-\eta \mathbf{H}_m)^t \cdot \mathbf{B}_0 \cdot(\mathbf{I}-\eta \mathbf{H}_m)^t  \rangle \\ \notag
    = &  \underbrace{\frac{\beta_m}{4} \operatorname{tr} ( (\mathbf{I}-(\mathbf{I}-\eta \mathbf{H}_m)^{N / 2} ) \mathbf{B}_0 ) \cdot  \langle \prod_{p=1}^{m-1} (\mathbf{I}-\eta \mathbf{H}_p )^{2N}  \mathbf{H}_m,  (\mathbf{I}-(\mathbf{I}-\eta \mathbf{H}_m)^{N / 2} )  \rangle}_{\text{bias term ${b_1^m}$}} \\ \notag
    + & \underbrace{ \frac{1}{2\eta} \langle \prod_{p=1}^{m-1} (\mathbf{I}-\eta \mathbf{H}_p )^{2N}  \cdot   (\mathbf{I}-(\mathbf{I}-\eta \mathbf{H}_m)^{2N} ) ,  \mathbf{B}_0  \rangle }_{\text{bias term $b_2^m$}}
\end{align}

The first bias item is lower bounded by:
$$
{\text{bias term ${b_1^m}$}}=\frac{\beta_m}{4} (\sum_i (1- (1-\eta \lambda_m^i )^{N / 2} ) \omega_i^2 ) \cdot (\sum_i \prod_{p=1}^{m-1}(1-\eta \lambda_p^i)^{2N} \lambda_m^i  (1- (1-\eta \lambda_m^i )^{N / 2} )  ),
$$
The second bias item is lower bounded by:
$$
{\text{bias term $b_2^m$}} \geq  (\sum_i \prod_{p=1}^{m-1}(1-\eta \lambda_p^i)^{2N}  (1- (1-\eta \lambda_m^i )^{2N } )\omega_i^2 )
$$
To further lower bound the two terms, we notice that:
$$
1- (1-\eta {\lambda_m^i} )^{\frac{N}{2}} \geq \begin{cases}1- (1-\frac{1}{N} )^{\frac{N}{2}} \geq 1-e^{-\frac{1}{2}} \geq \frac{1}{5}, & {\lambda_m^i} \geq \frac{1}{\eta N} \\ \frac{N}{2} \cdot \eta {\lambda_m^i}-\frac{N(N-2)}{8} \cdot \eta^2 {\lambda_m^i}^2 \geq \frac{N}{5} \cdot \eta {\lambda_m^i}, & {\lambda_m^i}<\frac{1}{\eta N}\end{cases}
$$
Substituting to the previous results, we have:
\begin{align} \notag
    {\text{bias term ${b_1^m}$}} &\geq \frac{\beta_m}{4} (\frac{1}{5} \cdot \sum_{i \leq {k_m^*}}  \omega_i^2+\frac{\eta N}{5} \sum_{i>{k_m^*}} (\lambda_m^i) \omega_i^2 )   \cdot  (\frac{1}{5} \cdot \sum_{i \leq {k_m^*}} \Gamma_{(1,m-1)}^i \lambda_m^i +\frac{\eta N}{5} \sum_{i>{k_m^*}} \Gamma_{(1,m-1)}^i (\lambda_m^i)^2 ) \\ \notag
    & = \frac{\beta_m}{25} \cdot  (  \|\mathbf{w}_0-\mathbf{w}^* \|_{ \mathbf{I}_{m,{0: {k_m^*}}}}^2 + N \eta  \|\mathbf{w}_0-\mathbf{w}^* \|_{\mathbf{H}_{m,{{k_m^*}:\infty }}}^2  )  \cdot  ( \sum_{i \leq {k_m^*}}{\Gamma_{(1,m-1)}^i }(\lambda_m^i)  +{\eta N} \sum_{i>{k_m^*}} \Gamma_{(1,m-1)}^i (\lambda_m^i)^2 ) 
\end{align}
and
\begin{align}\notag
    {\text{bias term $b_2^m$}} &\geq  (\frac{1}{5} \cdot \sum_{i \leq {k_m^*}} \Gamma_{(1,m-1)}^i \omega_i^2+\frac{\eta N}{5} \sum_{i>{k_m^*}} \Gamma_{(1,m-1)}^i \lambda_m^i \omega_i^2 ) \\ \notag
    & = \frac{1}{5} \cdot  ( \|\mathbf{w}_0-\mathbf{w}^* \|_{(\prod_{p=1}^{m-1} (\mathbf{I}-\eta \mathbf{H}_p )^{2N}  )_{0: {k_m^*}}}^2 + N \eta  \|\mathbf{w}_0-\mathbf{w}^* \|_{(\prod_{p=1}^{m-1} (\mathbf{I}-\eta \mathbf{H}_p )^{2N}  \mathbf{H}_m)_{{k_m^*}:\infty }}^2  )
\end{align}

Now we are ready to examine term 2.
\begin{align}\notag
    \text{term 2} =& \langle \hat{\Phi}_1^{m-1}\mathbf{I}, \sum_{t=0}^{N-1}  (\mathcal{I}-\eta {\mathcal{T}}_{\mathbf{H}_m} (\eta  ) )^t \mathbf{B}_{0}  \rangle \\ \notag 
    \geq &  \langle \hat{\Phi}_1^{m-1}\mathbf{I},  \frac{\beta_m}{4} \operatorname{tr} ( (\mathbf{I}-(\mathbf{I}-\eta \mathbf{H}_m)^{N / 2} ) \mathbf{B}_0 ) \cdot (\mathbf{I}-(\mathbf{I}-\eta \mathbf{H}_m)^{N / 2} )  \rangle \\ \notag
    + &   \langle \hat{\Phi}_1^{m-1}\mathbf{I},  \sum_{t=0}^{N-1}(\mathbf{I}-\eta \mathbf{H}_m)^t \cdot \mathbf{B}_0 \cdot(\mathbf{I}-\eta \mathbf{H}_m)^t  \rangle \\ \notag
    = &  \underbrace{\frac{\beta_m}{4} \operatorname{tr} ( (\mathbf{I}-(\mathbf{I}-\eta \mathbf{H}_m)^{N / 2} ) \mathbf{B}_0 ) \cdot  \langle \hat{\Phi}_1^{m-1}\mathbf{I},  (\mathbf{I}-(\mathbf{I}-\eta \mathbf{H}_m)^{N / 2} )  \rangle}_{\text{bias term ${d_1^m}$}} \\ \notag
    + & \underbrace{ \frac{1}{2\eta} \langle \hat{\Phi}_1^{m-1}\mathbf{H}_m^{-1}  \cdot   (\mathbf{I}-(\mathbf{I}-\eta \mathbf{H}_m)^{2N} ) ,  \mathbf{B}_0  \rangle }_{\text{bias term $d_2^m$}}
\end{align}
Analogous to term 1, we have:
\begin{align*}
    {\text{bias term ${d_1^m}$}} &=\frac{\beta_m}{4} (\sum_i (1- (1-\eta \lambda_m^i )^{N / 2} ) \omega_i^2 ) \cdot (\sum_i \hat{\Phi}_1^{m-1} \lambda_m^i  (1- (1-\eta \lambda_m^i )^{N / 2} )  ) \\
    & \geq \frac{\beta_m}{25} \cdot  (  \|\mathbf{w}_0-\mathbf{w}^* \|_{ \mathbf{I}_{m,{0: {k_m^*}}}}^2 + N \eta  \|\mathbf{w}_0-\mathbf{w}^* \|_{\mathbf{H}_{m,{{k_m^*}:\infty }}}^2  )  \cdot {\Phi_1^{m-1}} \cdot ( \sum_{i<{k_m^*}}(\lambda_m^i)   +{\eta N} \sum_{i>{k_m^*}}  (\lambda_m^i)^2 ) 
\end{align*}
and
\begin{align*}
    {\text{bias term $d_2^m$}} &\geq  (\sum_i \hat{\Phi}_1^{m-1} (\lambda_m^i)^{-1}(1- (1-\eta \lambda_m^i )^{2N } )\omega_i^2 )\\
    &\geq\hat{\Phi}_1^{m-1}  (\frac{1}{5} \cdot \sum_{i \leq {k_m^*}}  (\lambda_m^i)^{-1} \omega_i^2+\frac{\eta N}{5} \sum_{i>{k_m^*}}   \omega_i^2 ) \\ \notag
    & = \frac{\hat{\Phi}_1^{m-1}}{5} \cdot  ( \|\mathbf{w}_0-\mathbf{w}^* \|_{( \mathbf{H}_m^{-1})_{0: {k_m^*}}}^2 + N \eta \cdot  \|\mathbf{w}_0-\mathbf{w}^* \|_{ \mathbf{I})_{{k_m^*}:\infty }}^2  )
\end{align*}
After $MN$ iterations, it holds that:
\begin{align}\notag
    \mathbf{B}_{MN} & \succeq \prod_{m=1}^{M}  (\mathcal{I}-\eta \widetilde{\mathcal{T}}_{\mathbf{H}_m} (\eta  ) )^N \circ \mathbf{B}_{0} + \sum_{m=1}^{M}\prod_{j=m}^M  (\mathcal{I}-\eta \widetilde{\mathcal{T}}_{\mathbf{H}_j} (\eta  ) )^N \mathbf{P}_m,
\end{align}
where denoting $\mathbf{P}_m =\beta_m\eta^2 (b_1^m + b_2^m+d_1^m + d_2^m)\cdot  (\mathbf{I}-\eta \mathbf{H}_m )^{2N} \mathbf{H}_m  $.

Then, the bias error can be represented as follows:
\begin{align} \notag
     \sum_{k=1}^{M}  \langle \mathbf{H}_k, \mathbf{B}_{MN}  \rangle  &\geq \underbrace{\sum_{k=1}^{M}   \langle \mathbf{H}_k, \prod_{m=1}^{M}  (\mathcal{I}-\eta \widetilde{\mathcal{T}}_{\mathbf{H}_m} (\eta  ) )^N \circ \mathbf{B}_{0}  \rangle}_{\text{bias term 1'}}  + \underbrace{\sum_{k=1}^{M}  \langle \mathbf{H}_k, \sum_{m=1}^{M}\prod_{j=m}^M  (\mathcal{I}-\eta \widetilde{\mathcal{T}}_{\mathbf{H}_j} (\eta  ) )^N \mathbf{P}_m  \rangle}_{\text{bias term 2'}}. \\ \notag
     & \geq \sum_{k=1}^{M}   \|\mathbf{w}_0-\mathbf{w}^* \|_{\prod_{m=1}^{M}  (\mathbf{I}-\eta {\mathbf{H}_m} )^{2N}\mathbf{H}_k}^2 \\ \notag
     & + \sum_{k=1}^{M}  \langle \mathbf{H}_k, \sum_{m=1}^{M}\prod_{j=m}^M  (\mathbf{I}-\eta \mathbf{H}_j )^{2N} \beta_m\eta^2 (b_1^m + b_2^m+d_1^m + d_2^m)\cdot  (\mathbf{I}-\eta \mathbf{H}_m )^{2N} \mathbf{H}_m  \rangle.    
\end{align}
It follows that:
\begin{align*}
    \sum_{k=1}^{M}  \langle \mathbf{H}_k, \mathbf{B}_{MN}  \rangle & \geq \sum_{k=1}^{M}   \|\mathbf{w}_0-\mathbf{w}^* \|_{\prod_{m=1}^{M}  (\mathbf{I}-\eta {\mathbf{H}_m} )^{2N}\mathbf{H}_k}^2 \\ \notag
     & + \sum_{k=1}^{M} \sum_i \lambda_k^i \cdot \sum_{m=1}^{M}\prod_{j=m}^M  (1-\eta \lambda_j^i )^{2N} \beta_m\eta^2 (b_1^m + b_2^m+d_1^m + d_2^m)\cdot  (1-\eta \lambda_m^i )^{2N} \lambda_m^i  \\ \notag
     & \geq \sum_{k=1}^{M}   \|\mathbf{w}_0-\mathbf{w}^* \|_{\prod_{m=1}^{M}  (\mathbf{I}-\eta {\mathbf{H}_m} )^{2N}\mathbf{H}_k}^2 \\ \notag
     & + \sum_{k=1}^{M}  \sum_{m=1}^{M}({b_1^m}^{\prime} + {b_2^m}^{\prime} +{d_1^m}^{\prime} + {d_2^m}^{\prime})
\end{align*}
where
\begin{align} \notag
    {b_1^m}^{\prime} =& \frac{\beta_m^2 \eta^2}{25} \cdot  (  \|\mathbf{w}_0-\mathbf{w}^* \|_{ \mathbf{I}_{m,{0: {k_m^*}}}}^2 + N \eta  \|\mathbf{w}_0-\mathbf{w}^* \|_{\mathbf{H}_{m,{{k_m^*}:\infty }}}^2  ) 
    \\&\cdot  ( \sum_{i<k_m^*} {\Gamma_{(1,M)}^i \lambda_k^i (\lambda_m^i)^2}   +{\eta N} \sum_{i>k_m^*}  \nonumber\Gamma_{(1,M)}^i(\lambda_m^i)^3\lambda_k^i  )\\\notag
    {b_2^m}^{\prime} =& \frac{\beta_m \eta^2}{5} \cdot  (\|\mathbf{w}_0-\mathbf{w}^* \|_{(\bm{\Gamma}_{(1,M)} \mathbf{H}_m \mathbf{H}_k)_{0: {k_m^*}}}^2+N \eta  \|\mathbf{w}_0-\mathbf{w}^* \|_{(\bm{\Gamma}_{(1,M)}  (\mathbf{I}-\eta \mathbf{H}_m)^{2N} 
  \mathbf{H}_m^2 \mathbf{H}_k)_{{k_m^*}:\infty }}^2),
\end{align}
and
\begin{align*}
    {d_1^m}^{\prime} =& \frac{\beta_m}{25} \cdot  (  \|\mathbf{w}_0-\mathbf{w}^* \|_{ \mathbf{I}_{m,{0: {k_m^*}}}}^2 + N \eta  \|\mathbf{w}_0-\mathbf{w}^* \|_{\mathbf{H}_{m,{{k_m^*}:\infty }}}^2  ) 
    \\&\cdot \hat{\Phi}_1^{m-1} \cdot( \sum_{i<k_m^*} {\Gamma_{(m,M)}^i \lambda_k^i (\lambda_m^i) } +{\eta N} \sum_{i>k_m^*} \Gamma_{(m,M)}^i(\lambda_m^i)^2 \lambda_k^i )\\\notag
    {d_2^m}^{\prime} =& \frac{\beta_m \eta^2 \hat{\Phi}_1^{m-1}}{5} \cdot  (\|\mathbf{w}_0-\mathbf{w}^* \|_{(\bm{\Gamma}_{(m,M)} \mathbf{H}_k)_{0: {k_m^*}}}^2+N \eta  \|\mathbf{w}_0-\mathbf{w}^* \|_{(\bm{\Gamma}_{(m,M)}  (\mathbf{I}-\eta \mathbf{H}_m)^{2N} 
  \mathbf{H}_m \mathbf{H}_k)_{{k_m^*}:\infty }}^2).
\end{align*}
\section{Extension work}\label{app:extension}

It is noticed that when the step size is set to $\|\mathbf{x}_m\|^{-2}$, the update rule for the minimum norm solution can be considered equivalent to that of the last iterate SGD. Consequently, in this subsection, we will focus on a particular case (akin to the setting in \citealt{lin2023theory}) that involves this specific step size, allowing us to draw direct comparisons and insights under a defined set of conditions.

Consider a series of tasks $\mathbb{M}=\{1,2, \ldots, M\}$. Given $M$ datasets, for each dataset $m \in \mathbb{M}$, $D_m = \{ (\mathbf{x}_{m,i},y_{m,i} ) \}_{i=1}^{N}$ drawn i.i.d from 
some fixed distribution $\mathcal{D}_m=\mathcal{X}_m \times \mathcal{Y}_m \subset \mathbb{R}^d \times \mathbb{R}$. Assume that $ \{(\mathbf{x}_{m,i}, y_{m,i}) \}_{i=1}^N$ are i.i.d. sampled from a linear regression model, i.e., each $(\mathbf{x}_{m,i}, y_{m,i})$ is a realization of the linear regression model $ y_m  =  (\mathbf{x}_m^{\top} \mathbf{w}_m^* ) + z_m$, where $z_m$ is some randomized noise satisfing well-specified condition and $\mathbf{w}_m^*\in \mathbb{R}^d$ is the optimal model parameter for task $m$. 

We adopt the same learning procedure with specific step size, aiming to output a model $\mathbf{w}_M^N$ minimizing the \emph{performance} \cite{lin2023theory}, i.e.
\begin{equation}
    G(\mathbf{w}_M^N)=\frac{1}{M} \sum_{i=1}^M \|\mathbf{w}_M^N-\mathbf{w}_i^* \|^2.
\end{equation}
Therefore, our results can be restated as follows
\begin{theorem}\label{thm:specific_case}
Consider a scenario where the model $\mathbf{w}$ undergoes training via SGD for $M$ distinct tasks, following a sequence $1, \ldots, M$. With a specific step size of $\eta_{m,t}=\|\mathbf{x}_{m,t}\|^{-2}$, each task is executed for $N$ iterations. Given that Assumption \ref{asm:nosie} are satisfied, the following will hold:
    {\small
    \begin{equation*}
   \begin{aligned}
        \mathbb{E}  [G (\mathbf{w}_M^N ) ] & =  \frac{1}{M} \sum_{i=1}^{M}\|\mathbf{w}_0^{0} - \mathbf{w}_i^* \|_{\prod_{m=1}^M\prod_{t=1}^N\left(\mathbf{I}-{\mathbf{H}_m}{\eta_{m,t}}\right)}^2 \\
    & +  \frac{1}{M}\sum_{i=1}^{M}\sum_{m=1}^{M}  \sum_{t=0}^{N-1} \| \mathbf{w}_m^* -\mathbf{w}_i^*\|_{\prod_{p=1}^{M-m}\prod_{j=1}^N \left(\mathbf{I}-{\mathbf{H}_p}{\eta_{p,j}}\right) \prod_{j=q}^{N-t}\left(\mathbf{I}-{\mathbf{H}_m}{\eta_{m,q}}\right) {\mathbf{H}_m}{\eta_{m,q}}}^2   \\
    & + \frac{1}{M} \sum_{i=1}^{M}\sum_{m=1}^{M}  \sum_{t=0}^{N-1} \| \bm{z}_{m,t}\|_{\prod_{p=1}^{M-m}\prod_{j=1}^N \left(\mathbf{I}-{\mathbf{H}_p}{\eta_{p,j}}\right) \prod_{q=1}^{N-t}\left(\mathbf{I}-{\mathbf{H}_m}{\eta_{m,q}}\right) {\eta_{m,q}}}^2. 
   \end{aligned}
\end{equation*}}
\end{theorem}
\begin{remark}
    In contrast to the approach in \cref{thm:main_up} and \cref{thm:main_low}, here we do not rely on the decomposition of bias and variance error while considering that the projection $(\mathbf{I}-\eta_{m,t} \mathbf{x}_{m,t}\mathbf{x}_{m,t}^{\top})$ is orthogonal to $\eta_{m,t} \mathbf{x}_{m,t}\mathbf{x}_{m,t}^{\top}$ with a specific stepsize $\eta_{m,t}=\|\mathbf{x}_{m,t}\|_2^{-2}$. This perspective allows us to derive a closed-form expression for the expected performance, which integrates the impact of initial parameter deviations, task-specific parameter variations, and random noise. 
    Furthermore, \cref{thm:specific_case} in our study explores the performance behavior on general data distributions, expanding beyond the Gaussian distribution context discussed in \citealt{lin2023theory}. In scenarios where there is only a single sample per training iteration, our results could cover their findings. 
\end{remark}

\begin{proof}[{\bf Proof}]
    For each iteration, according to the update rule of SGD, it holds that
    $$
    \mathbf{w}_m^N = \mathbf{w}_m^{N-1}- \eta (\bm{x}_{m,N}( (\bm{x}_{m,N})^{\top}\mathbf{w}_m^{N-1} -y_{m,N} ) ).
    $$
    which can be rewritten as:
    $$
    \mathbf{w}_m^N - \mathbf{w}_i^* = (\mathbf{I}- \eta\bm{x}_{m,N} (\bm{x}_{m,N})^{\top}) (\mathbf{w}_m^{N-1}- \mathbf{w}_i^*) + \eta z_{m,N} \bm{x}_{m,N}.
    $$

    We consider the expectation norm for both sides:
$$
\begin{aligned}
    &\mathbb{E}  [ \|\mathbf{w}_m^N - \mathbf{w}_i^* \|^2 ]  \\
    =&\mathbb{E}  [  ( \mathbf{w}_m^N - \mathbf{w}_i^*  )^{\top} ( \mathbf{w}_m^N - \mathbf{w}_i^*  ) ] \\
     =&\mathbb{E}  [  ( \mathbf{w}_m^{N-1} - \mathbf{w}_i^*  )^{\top}  ( \mathbf{I}- \eta\bm{x}_{m,N} (\bm{x}_{m,N})^{\top} )^{\top} ( \mathbf{I}- \eta\bm{x}_{m,N} (\bm{x}_{m,N})^{\top} ) ( \mathbf{w}_m^{N-1} - \mathbf{w}_i^*  )+ \eta^2  (z_{m,N} \bm{x}_{m,N} )^{\top} (z_{m,N} \bm{x}_{m,N} ) ]\\
     (*)= &\mathbb{E}  [ \|  (\mathbf{I}-  \eta_{m,N}\bm{x}_{m,N} (\bm{x}_{m,N})^{\top} )  (\mathbf{w}_m^{N-1}- \mathbf{w}_i^* )\|^2 + \| \eta_{m,N}\bm{x}_{m,N} (\bm{x}_{m,N})^{\top}  (\mathbf{w}_m^* -\mathbf{w}_i^*  ) \|^2+ \|  \eta_{m,N} \bm{x}_{m,N} \bm{z}_{m,N}\|^2  ] \\
    = &  \mathbb{E}  [ \|\mathbf{w}_m^{N-1} - \mathbf{w}_i^* \|_{ (\mathbf{I}-{\mathbf{H}_m}{\eta_{m, N}} )}^2 ]+ {\eta_{m,N}} \sigma^2 + \| \mathbf{w}_m^* -\mathbf{w}_i^*\|_{{\mathbf{H}_m}{\eta_{m,N}}}^2\\
    = &  \|\mathbf{w}_m^{0} - \mathbf{w}_i^* \|_{\prod_{t=1}^N (\mathbf{I}-{\mathbf{H}_m}{\eta_{m,t}} )}^2 
    +  \sum_{t=1}^{N-1} \| \mathbf{w}_m^* -\mathbf{w}_i^*\|_{\prod_{j=1}^{N-t}  (\mathbf{I}-{\mathbf{H}_m}{\eta_{m,j}} )^j {\mathbf{H}_m}{\eta_{m,N}}}^2  
    + \sum_{t=1}^{N-1} \| \bm{z}_{m,t}\|_{\prod_{j=1}^{N-t}  (\mathbf{I}-{\mathbf{H}_m}{\eta_{m,j}} )^j {\eta_{m,N}}}^2, 
\end{aligned}
$$
where the (*) equation comes from the choice of step size such that $ (\mathbf{I}-  \eta_{m,t}\bm{x}_{m,t} (\bm{x}_{m,t})^{\top} )$ and $ \eta_{m,t}\bm{x}_{m,t} (\bm{x}_{m,t})^{\top} $ are orthogonal projection, which equals the minimum norm solution with one sample.

Considering $M$ tasks, it holds that
\begin{equation*}
    \begin{aligned}
    \mathbb{E} \|\mathbf{w}_{m,} - \mathbf{w}_i^*\|^2 & =  \|\mathbf{w}_0^{0} - \mathbf{w}_i^* \|_{\prod_{m=1}^M\prod_{t=1}^N(\mathbf{I}-{\mathbf{H}_m}{\eta_{m,t}})}^2 \\
    & +  \sum_{m=1}^{M-1}  \sum_{t=0}^{N-1} \| \mathbf{w}_m^* -\mathbf{w}_i^*\|_{\prod_{p=1}^{M-m}\prod_{j=1}^N (\mathbf{I}-{\mathbf{H}_p}{\eta_{p,j}}) \prod_{j=q}^{N-t}(\mathbf{I}-{\mathbf{H}_m}{\eta_{m,q}}) {\mathbf{H}_m}{\eta_{m,q}}}^2   \\
    & +\sum_{m=1}^{M-1}  \sum_{t=0}^{N-1} \| \bm{z}_m\|_{\prod_{p=1}^{M-m}\prod_{j=1}^N (\mathbf{I}-{\mathbf{H}_p}{\eta_{p,j}}) \prod_{q=1}^{N-t}(\mathbf{I}-{\mathbf{H}_m}{\eta_{m,q}}) {\eta_{m,q}}}^2. 
\end{aligned}
\end{equation*}

In conclusion, we aggregate the performance metrics across tasks, ranging from $i=1$ to $i=M$, to derive the final result.
\end{proof}


\end{document}